\documentclass{article} 
\usepackage{iclr2027_conference,times}
\iclrfinalcopy  

\usepackage{amsmath,amsfonts,bm}

\def\eqref#1{equation~\ref{#1}}

\def\1{\bm{1}}

\DeclareMathAlphabet{\mathsfit}{\encodingdefault}{\sfdefault}{m}{sl}
\SetMathAlphabet{\mathsfit}{bold}{\encodingdefault}{\sfdefault}{bx}{n}

\usepackage{url}
\usepackage{algorithm}
\usepackage{algpseudocode}
\usepackage{amsmath}  
\usepackage{amssymb}
\usepackage{multirow}
\usepackage{float}
\usepackage{wrapfig}
\usepackage{caption}
\usepackage{graphicx}
\usepackage{subcaption}
\usepackage{booktabs}
\usepackage{bbm}
\usepackage{hyperref}

\title{ERASE: Eliminating Redundant Visual Tokens via Adaptive Two-Stage Token Pruning}

\author{%
  Yuna Lee$^{1}$ \qquad Kyoungho Min$^{1}$ \qquad Yulhwa Kim$^{2,}$\thanks{Corresponding author} \\[0.5em]
  $^{1}$Department of Electrical and Computer Engineering, Sungkyunkwan University, Republic of Korea \\
  $^{2}$Department of Semiconductor Systems Engineering, Sungkyunkwan University, Republic of Korea \\[0.5em]
  \texttt{\{yuna6548, kyohmin, yulhwakim\}@skku.edu} \\[0.2em]
}

\begin{document}

\maketitle
\fancyhead[L]{Preprint}
\begin{abstract}
    Recent advances in Vision-Language Models (VLMs) enable large language models (LLMs) to process high-resolution images, significantly improving real-world multimodal understanding. However, this capability introduces a large number of vision tokens, incurring substantial computational overhead. To mitigate this issue, various vision token compression methods have been proposed. 
    Existing methods often estimate different sources of visual redundancy using learned representations or fixed pruning schedules. We propose ERASE, an adaptive two-stage framework that separates image-level redundancy removal from instruction-dependent token pruning.
    Stage 1 derives image-dependent token retention from lightweight raw-image statistics, while Stage 2 progressively removes instruction-irrelevant tokens across decoder layers.
    Experiments demonstrate substantial token reduction while preserving accuracy: on Qwen2.5-VL-7B, ERASE retains 95.70\% of the original model's accuracy at 25\% token retention. Our code is available at \url{https://github.com/Tuna-Luna/ERASE}.
\end{abstract}

\section{Introduction}
\label{sec:intro}

Recent VLMs augment LLMs with visual understanding capabilities by jointly processing text and image modalities~\citep{Qwen2_5_VL, chen2024internvl}. However, visual inputs typically generate a large number of vision tokens, significantly increasing computational cost and inference latency. Vision token compression mitigates this bottleneck by removing redundant or less relevant tokens~\citep{wen2025stop, zhang2025cdpruner, fang2026prunesid, fastv, baek2026agilepruner, zhang2026vscan, huang2026nwa, hu2026illava}. 
Meanwhile, existing approaches assess visual redundancy using learned semantic features or identify instruction-relevant tokens within the LLM, often overlooking the potential of image-adaptive retention guided by raw-image statistics and layer-wise changes in text–vision alignment.

In this paper, we propose \textit{ERASE: Eliminating Redundant visual tokens via Adaptive two-StagE token pruning}, a hierarchical vision token pruning framework that compresses vision tokens through two complementary pruning stages. ERASE decomposes vision-token redundancy into two complementary sources and removes them at different points in the VLM pipeline. 
Stage 1 adaptively removes image redundancy before the LLM using each image's own statistics, 
while Stage 2 removes instruction-irrelevant tokens within the decoder through hierarchical pruning. This separation avoids relying on expensive encoder attention maps for early redundancy removal while preserving instruction-dependent information later in the model.
Extensive experiments demonstrate that ERASE consistently achieves a superior efficiency-performance trade-off over existing vision token compression methods across diverse VLM architectures and benchmarks.

\section{Motivation}
\subsection{Vision token scaling in modern VLMs}
\label{sec:vision_tokens}
VLMs process images by partitioning them into patches and converting into visual tokens for LLM processing. Early VLMs used pretrained vision encoders, such as CLIP~\citep{clip}. These encoders typically operated at fixed input resolutions. For instance, the original LLaVA~\citep{llava} encodes $224\times224$ images into 256 visual tokens using CLIP. Resizing high-resolution images to a fixed resolution can discard fine-grained details and degrade visual understanding.
To address this limitation, advanced VLMs adopt dynamic tiling strategies that partition high-resolution images into multiple tiles~\citep{liu2024llavanext,wu2024deepseekvl2mixtureofexpertsvisionlanguagemodels,zhu2025internvl3}. For instance, InternVL3~\citep{zhu2025internvl3} selects a tile grid according to the image's aspect ratio and adds a global thumbnail to preserve global context. Recent models such as Qwen2.5-VL~\citep{Qwen2_5_VL} and Qwen3-VL~\citep{Qwen3-VL} further support fully configurable resolutions using dedicated image encoders.

\begin{figure}
  \centering
  \includegraphics[width=1.0\linewidth]{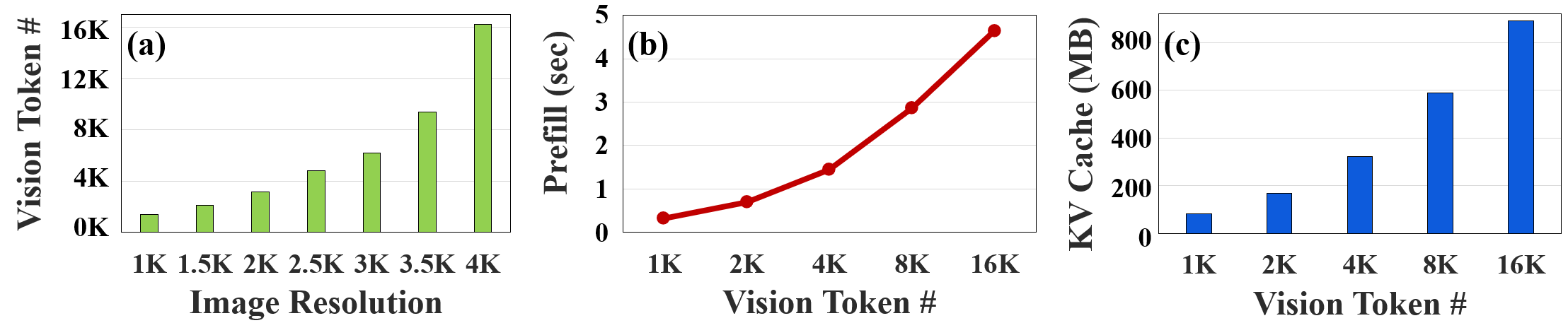}
  \vspace{-18pt}
  \caption{Scaling behavior of (a) vision token count with increasing image resolution, and (b) prefill latency and (c) KV cache size with increasing vision token count in Qwen2.5-VL-7B.
  }
  \label{fig:vision_token_overhead}
  \vspace{-12pt}
\end{figure}

However, recent VLMs that support high-resolution image processing substantially increase the number of vision tokens. Since images are represented as 2D spatial grids, the number of tokens scales quadratically with image resolution (Fig.~\ref{fig:vision_token_overhead}a). For example, a 2K image generates approximately 3K tokens, while a 4K image produces around 16K tokens. Attention complexity scales quadratically with sequence length, and the token explosion significantly increases prefill latency (Fig.~\ref{fig:vision_token_overhead}b) and KV cache memory usage (Fig.~\ref{fig:vision_token_overhead}c). 
Therefore, efficient VLM inference requires effective vision token compression.

\begin{figure}[t]
  \centering
  \begin{subfigure}[b]{0.32\textwidth} 
    \centering
    \includegraphics[width=\textwidth]{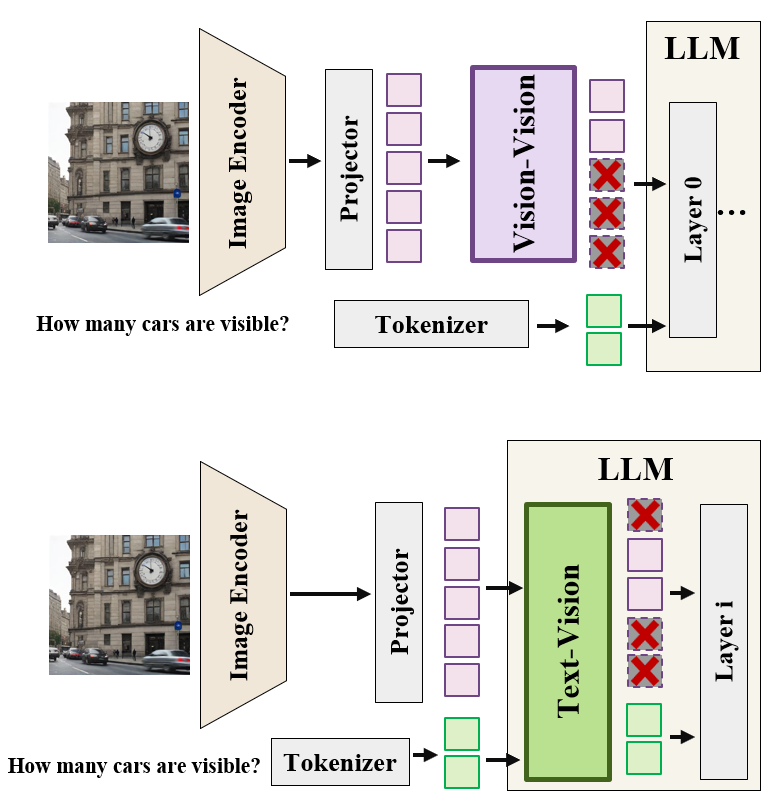} 
    \vspace{-12pt}
    \caption{Single-objective compression}
    \label{fig:methods-a}
  \end{subfigure}\hfill 
  \begin{subfigure}[b]{0.35\textwidth}
    \centering
    \includegraphics[width=\textwidth]{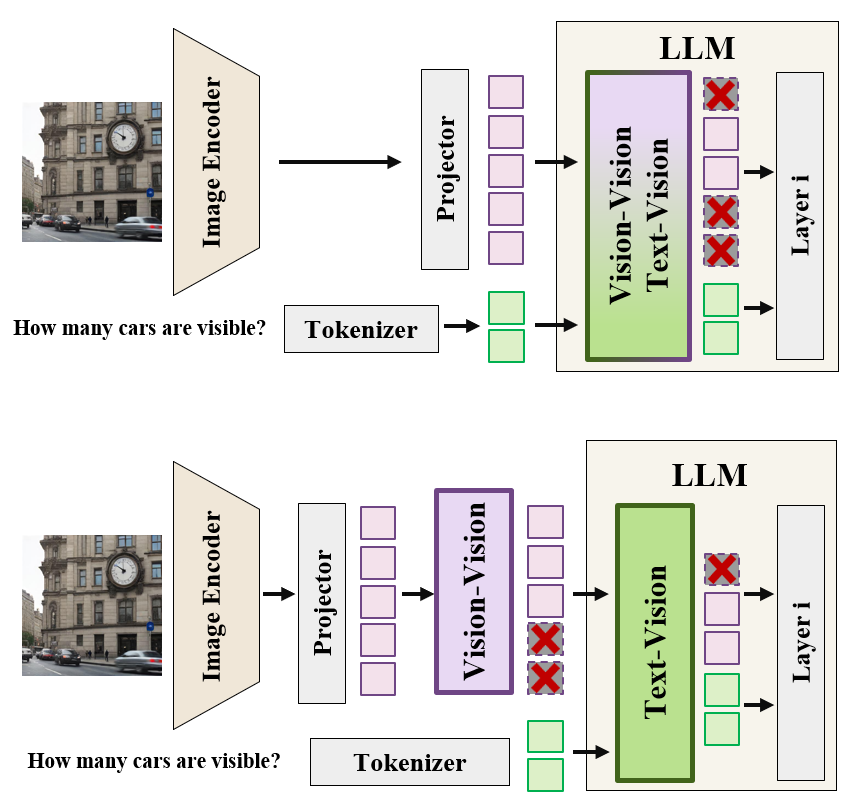}
    \vspace{-12pt}
    \caption{Hybrid compression}
    \label{fig:methods-b}
  \end{subfigure}\hfill 
  \begin{subfigure}[b]{0.28\textwidth}
    \centering
    \includegraphics[width=\textwidth]{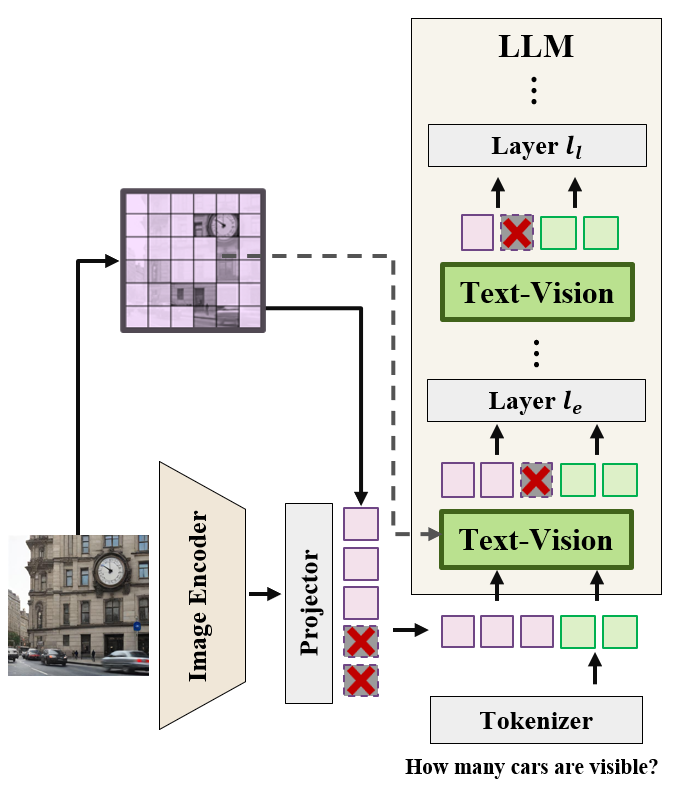}
    \vspace{-12pt}
    \caption{Proposed ERASE}
    \label{fig:methods-c}
  \end{subfigure}
  \vspace{-5pt}
  \caption{Comparison of vision token compression schemes. (a) (Top) vision-only compression~\citep{fang2026prunesid,baek2026agilepruner}; (Bottom) text-conditioned compression~\citep{fastv,Xing_2025_CVPR_Pdrop}. (b) (Top) single-stage hybrid compression~\citep{zhang2025cdpruner,wen2025stop}; (Bottom) multi-stage hybrid compression~\citep{zhang2026vscan,huang2026nwa,hu2026illava}. (c) image-adaptive two-stage hybrid compression.
}
  \label{fig:methods}
\end{figure}

\subsection{Prior vision token compression methods}
As shown in Fig.~\ref{fig:methods}, vision token compression methods can be categorized into single-objective and hybrid approaches. Single-objective approaches focus on either visual redundancy or text-vision relevance, whereas hybrid approaches combine both in single-stage or multi-stage frameworks.

\subsubsection{Single-objective compression methods}\label{sec:single-modal}
\paragraph{Vision-only compression}
Vision-only methods remove redundant tokens before the LLM using attention scores or feature similarity. 
Attention-based methods~\citep{baek2026agilepruner,yang2025visionzip} primarily target CLIP-based VLMs such as LLaVA, which inherit a pretrained CLIP encoder with a \texttt{[CLS]} token trained as a global image representation aligned with text in a shared embedding space.
These methods can exploit the \texttt{[CLS]}-to-patch attention already available in the pretrained CLIP encoder as an efficient pruning cue. 
However, unlike CLIP-based encoders, many modern VLMs use dedicated vision encoders without a \texttt{[CLS]}-like global summary token.
\begin{wrapfigure}{r}{0.3\textwidth} 
  \centering
  \includegraphics[width=\linewidth]{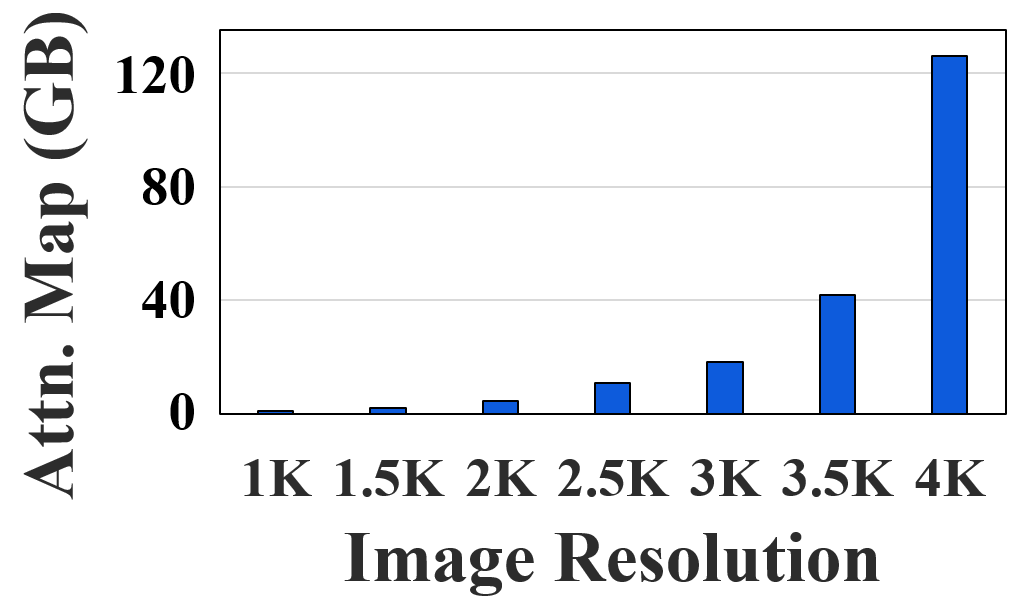} 
  \vspace{-20pt}
  \caption{
  Attention map scaling in the image encoder. 
  }
  \vspace{-12pt}
  \label{fig:vscan}
\end{wrapfigure}
A direct adaptation of such attention-based methods then requires access to the full $N\times N$ self-attention map. Moreover, image encoders often process more tokens than are passed to the LLM after vision-language projector downsampling. This causes prohibitive memory overhead for high-resolution inputs, exceeding 120~GB for Qwen2.5-VL-7B (Fig.~\ref{fig:vscan}) and making such methods impractical.
Feature-similarity-based methods instead group visually similar tokens and preserve those with lower similarity scores~\citep{fang2026prunesid}. However, visually important regions often depend on the text prompt, and ignoring text-vision relevance leaves room for further compression. Their effectiveness also depends heavily on the ability of the target model to encode semantic information into visual features.

\paragraph{Text-conditioned compression}
Text-conditioned approaches retain vision tokens that are highly relevant to the text instruction. These methods typically operate inside the LLM backbone, leveraging cross-attention to compress tokens receiving low attention weights from text tokens~\citep{fastv,Xing_2025_CVPR_Pdrop}. However, they overlook redundancy among similar vision tokens.
As noted by DART~\citep{wen2025stop}, once a representative token is retained from a group of similar tokens, the importance of the remaining redundant tokens should be adjusted.

\subsubsection{Hybrid compression strategies}
\paragraph{Single-stage hybrid compression}
Single-stage approaches jointly address visual redundancy and text relevance, typically at or just before the LLM decoder. 
For instance, DART~\citep{wen2025stop} employs similarity-based metrics using pivot tokens from both modalities, while CDPruner~\citep{zhang2025cdpruner} selects tokens before the LLM decoder by maximizing diversity conditioned on instruction relevance. However, these methods make compression decisions at a single point, without accounting for dynamic layer-wise evolution of vision-text relationships within the LLM backbone.

\paragraph{Multi-stage hybrid compression}
Multi-stage approaches generally combine both objectives by first reducing visual redundancy before the LLM decoder layers and then applying context-aware compression within it~\citep{zhang2026vscan,huang2026nwa,hu2026illava}. Separating these objectives can mitigate attention dilution among similar visual tokens before relevance filtering.
Despite this advantage, the methods adopted at each stage still present limitations. 
They often rely on attention in the image encoder to remove redundancy, introducing the same memory overhead discussed for vision-only methods.
VScan~\citep{zhang2026vscan} further exacerbates this issue as it requires two attention maps inside the image encoder, effectively doubling the severe memory overhead shown in Fig.~\ref{fig:vscan}. 
iLLaVA~\citep{hu2026illava} mitigates this overhead by accessing the softmax normalizer values that FlashAttention-2~\citep{dao2023flashattention2} computes, instead of materializing the full attention map. However, their redundancy estimates depend on learned visual representations, while stage-wise budgets remain fixed across images. This motivates model-independent early pruning with image-adaptive retention followed by context-aware pruning.

\section{Proposed method}
\subsection{Overview of ERASE}
\begin{figure}[t]
  \centering
  \includegraphics[width=1.0\linewidth]{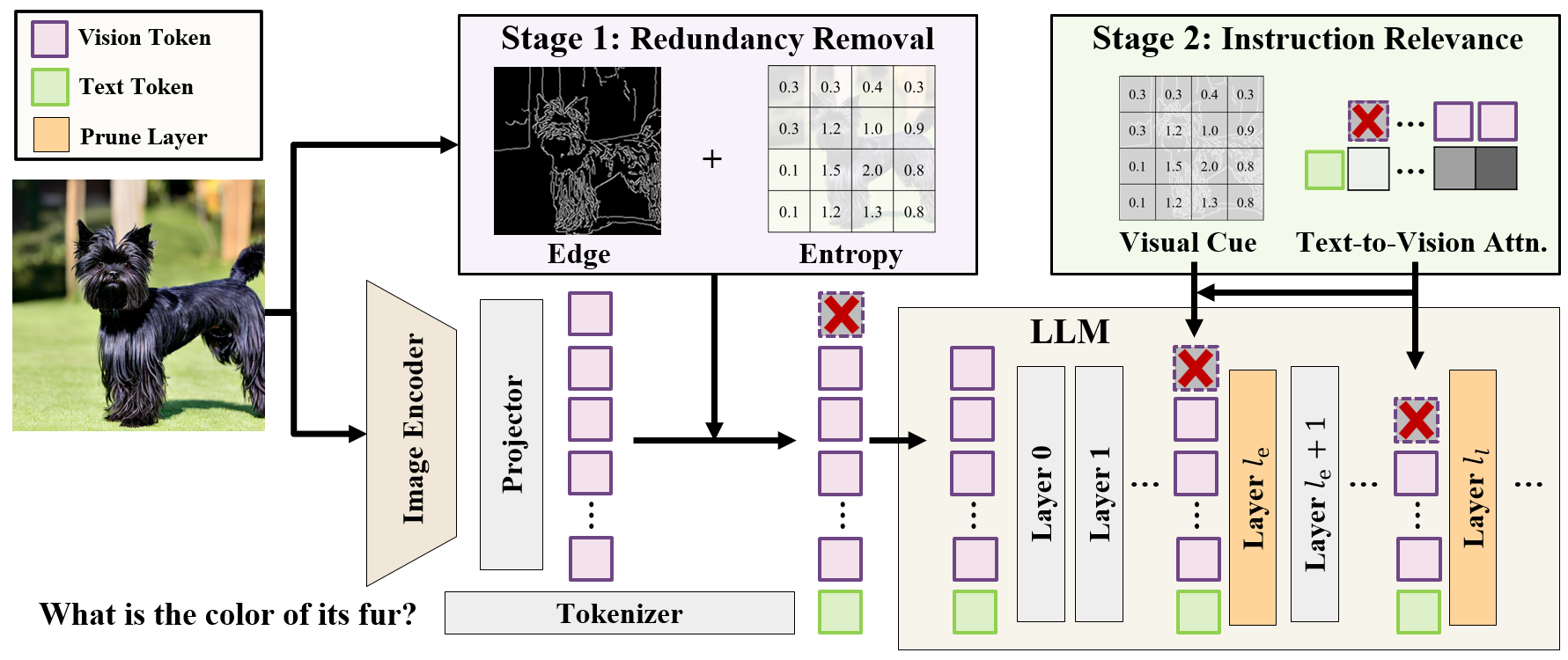}
  \vspace{-18pt}
  \caption{Overview of ERASE.}
  \vspace{-12pt}
  \label{fig:proposed}
\end{figure}

ERASE derives image-dependent token retention from raw-image statistics to reduce visual redundancy, followed by instruction-aware pruning (Fig.~\ref{fig:proposed}). Stage 1 jointly determines which visual tokens to retain and how many using image-relative redundancy cues. Stage 2 progressively removes instruction-irrelevant tokens at shallow and deep layers, reusing Stage 1 visual cues to complement shallow attention while relying on instruction relevance alone for deep pruning.

\subsection{Stage 1: Image-level vision token pruning}

Stage 1 exploits low-level image structure to identify visually redundant tokens, motivated by the well-established observation that structural information is concentrated around intensity transitions and high-frequency regions, while smooth regions are more spatially redundant. 
\begin{wrapfigure}{r}{0.35\textwidth} 
  \centering
  \includegraphics[width=\linewidth]{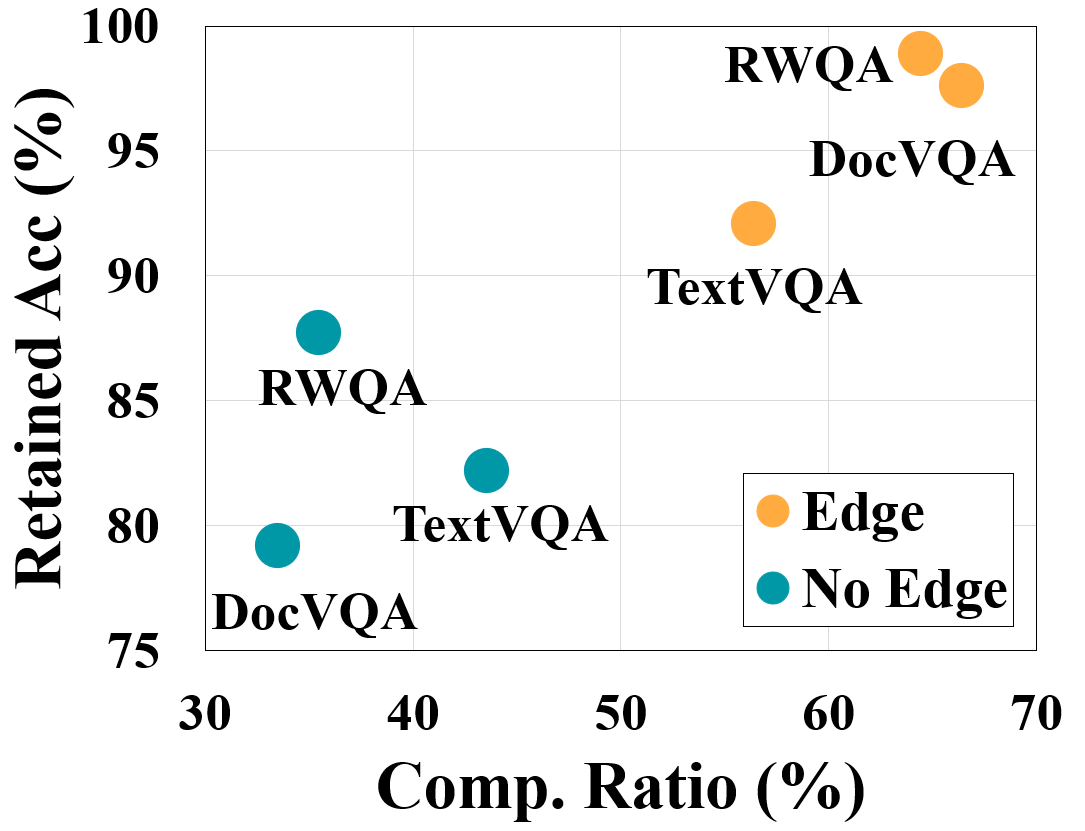} 
  \vspace{-18pt}
  \caption{
  Accuracy--compression of edge and non-edge patches.
  }
  \vspace{-10pt}
  \label{fig:edge_curve}
\end{wrapfigure}
We verify this for vision-token compression by separately retaining patches with and without edges. As shown in Fig.~\ref{fig:edge_curve}, edge-containing patches better preserve accuracy under compression, supporting edges as informative cues.
Based on this observation, ERASE uses edges as the primary cue for informative regions. To reduce sensitivity to color variation and local noise, the input is converted to luminance and smoothed by average pooling and Gaussian filtering, yielding $\tilde{Y}$, which is used only to derive pruning cues while the original image is passed unchanged to the image encoder. Edges from $\tilde{Y}$ capture structural boundaries but may miss informative regions with subtle variations. ERASE therefore complements them with patch-wise entropy to capture local information diversity. Together, the two cues identify informative patches and prune smooth, low-information regions.


\paragraph{Structural information estimation via edges} \label{sec:atp}

To detect edges, we compute the horizontal and vertical gradients of $\tilde{Y}$ using Sobel filters~\citep{sobel1968isotropic},
\begin{equation}
\label{eq:sobel}
G_x = S_x * \tilde{Y}, \qquad G_y = S_y * \tilde{Y},
\end{equation}
\begin{equation}
S_x =
\begin{bmatrix}
-1 & 0 & 1 \\
-2 & 0 & 2 \\
-1 & 0 & 1
\end{bmatrix}, \qquad
S_y =
\begin{bmatrix}
-1 & -2 & -1 \\
 0 &  0 &  0 \\
 1 &  2 &  1
\end{bmatrix},
\end{equation}
from which the gradient magnitude at pixel $(x,y)$ is computed as
\begin{equation}
\label{eq:grad_mag}
M(x,y) = \sqrt{G_x(x,y)^2 + G_y(x,y)^2}.
\end{equation}

\begin{figure}[t]
  \centering
  \begin{subfigure}[b]{0.25\textwidth} 
    \centering
    \includegraphics[width=\textwidth]{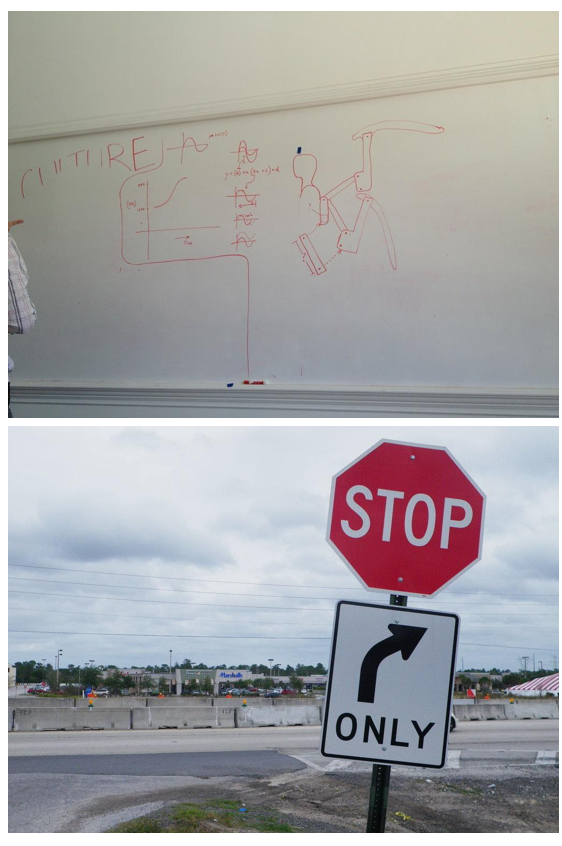} 
    \caption{Original patches} 
    \label{fig:stage1-a}
  \end{subfigure}
  \begin{subfigure}[b]{0.25\textwidth} 
    \centering
    \includegraphics[width=\textwidth]{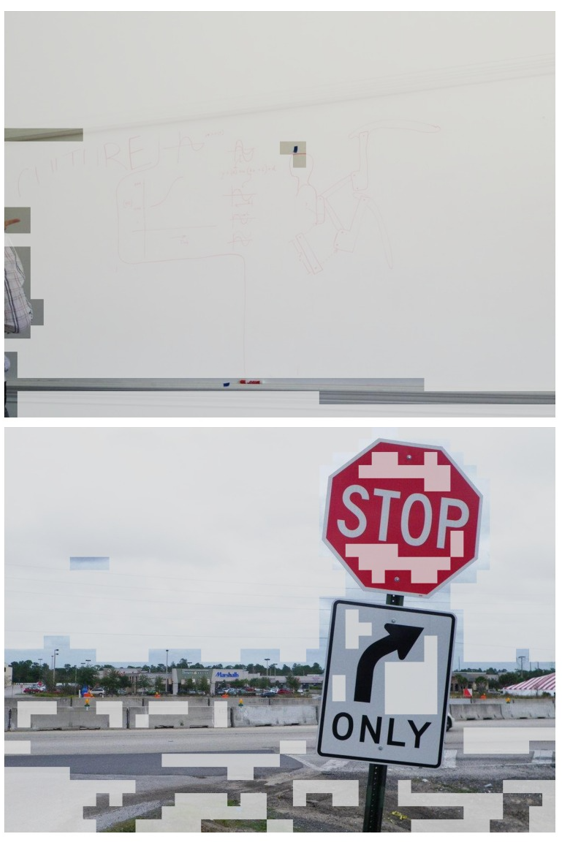} 
    \caption{ Edge patches
    } 
    \label{fig:stage1-b}
  \end{subfigure}
  \begin{subfigure}[b]{0.25\textwidth}
    \centering
    \includegraphics[width=\textwidth]{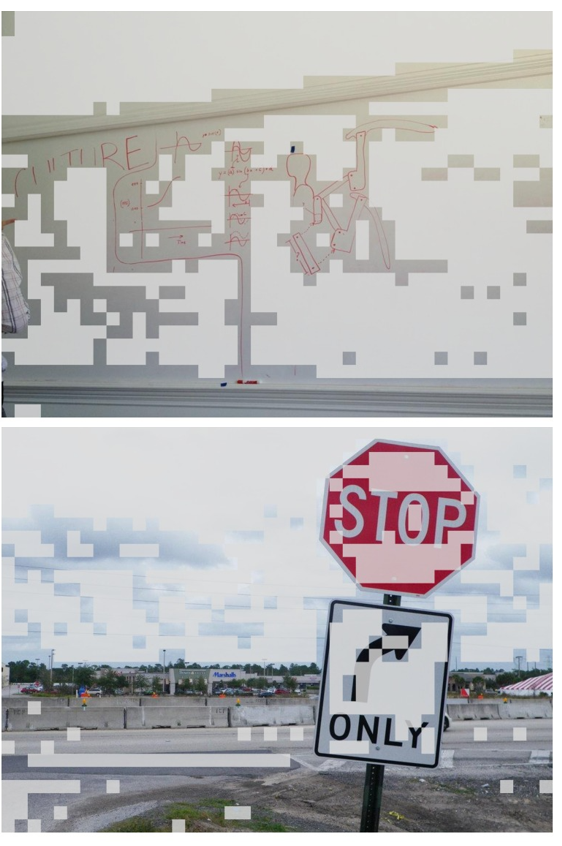}
    \caption{High-entropy patches}
    \label{fig:stage1-c}
  \end{subfigure}
  \begin{subfigure}[b]{0.25\textwidth}
    \centering
    \includegraphics[width=\textwidth]{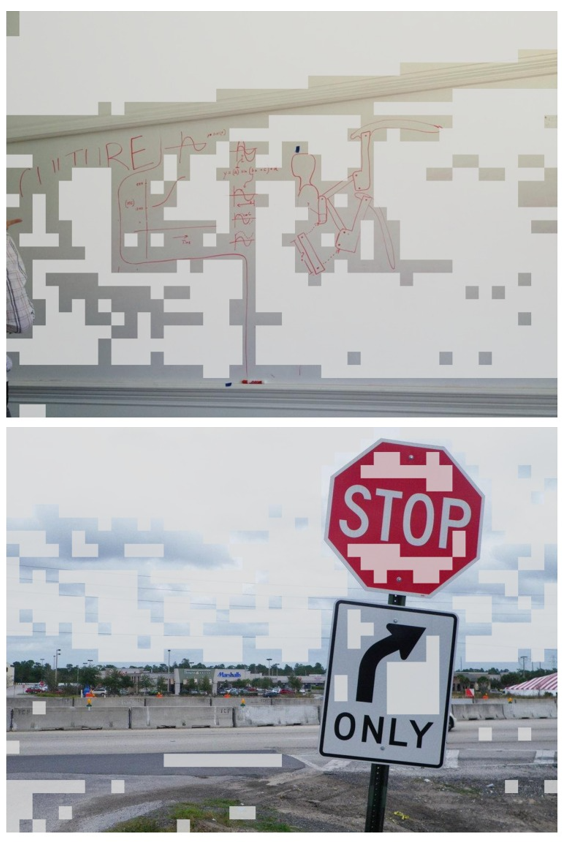}
    \caption{Combined patches}
    \label{fig:stage1-d}
  \end{subfigure}
  \caption{Retained regions selected by the edge, entropy, and combined criteria.}
  \label{fig:stage1}
\end{figure}

Following the Sobel gradient computation, we adopt Canny-inspired thresholding~\citep{canny1986}. Canny uses both high and low thresholds to preserve edge continuity. 
In contrast, we use a single adaptive threshold derived from the image median, since our goal is to identify informative regions rather than precisely trace edges.
With $\tau\in[0,1]$ controlling the edge sensitivity, we define
\begin{equation}
\label{eq:edge_rule}
\tau_E=(1-\tau)\,\mathrm{median}(\tilde{Y}),
\qquad
e(x,y)=\mathbf{1}\!\left[M(x,y)>\tau_E\right].
\end{equation}
A token is retained if its corresponding patch contains any detected edge pixel. Otherwise, the patch is treated as a candidate redundant region. Because gradient magnitude can miss informative low-contrast structures, faint handwriting on a bright board is almost entirely pruned in Fig.~\ref{fig:stage1-b} (top).
We address this limitation using the complementary entropy-based criterion described next.

\paragraph{Local information estimation via entropy}
To complement the edge-based criterion, we measure patch-wise intensity entropy from the preprocessed luminance map $\tilde{Y}$. While edge information captures spatial intensity changes, entropy captures the diversity of intensity values within each patch, providing a complementary cue for informative regions without strong boundaries.
For each patch $p$ of size $p_h\times p_w$, we construct a 256-bin histogram of its 8-bit luminance values and compute
\begin{equation}
\label{eq:patch_entropy}
H_p = -\sum_{v=0}^{255} \hat{P}_p(v) \ln \hat{P}_p(v),
\qquad
\hat{P}_p(v) = \frac{n_p(v)}{p_h \cdot p_w},
\end{equation}
where $n_p(v)$ is the number of pixels with luminance value $v$. Uniform patches concentrate their values in a few bins and thus have low entropy, whereas patches with diverse intensity values produce broader distributions and higher entropy. Since entropy distributions vary across images, ERASE uses the mean entropy across all patches, $\bar{H}$, as an image-adaptive threshold and retains a patch when $H_p>\bar{H}$. Unlike gradient magnitude, $H_p$ depends on the distribution of intensity values rather than their spatial differences, allowing it to remain high even when local variations are too subtle to form strong edges. As shown in Fig.~\ref{fig:stage1-c} (top), this enables the entropy criterion to retain faint handwriting that is discarded by the edge criterion.
However, entropy measures intensity diversity without accounting for spatial structure, so informative but locally uniform patches may fall below $\bar{H}$, such as the simple arrow preserved by edges in Fig.~\ref{fig:stage1} (bottom).
To preserve regions identified by either cue, ERASE prunes a token only when its patch contains no detected edge and $H_p\leq\bar{H}$ (Fig.~\ref{fig:stage1-d}).
This complementary rule preserves structural boundaries and local intensity diversity while adapting retention to each image's statistics rather than a prescribed token count.

\subsection{Stage 2: Context-aware vision token pruning}

\paragraph{Layer-wise vision--text dynamics}
Prior studies suggest hierarchical multimodal processing across decoder layers~\citep{wu2025accelerating_dyvte, shi2025vision_vfl}. DyVTE~\citep{wu2025accelerating_dyvte} identifies early fusion, intra-modality modeling, and multimodal reasoning, with cross-modal interaction decreasing after early fusion and increasing again in deeper layers. Likewise, 
~\citet{shi2025vision_vfl} associate shallow layers with basic visual recognition and deeper layers with fine-grained visual-semantic tasks.
\begin{figure}
  \centering
  \begin{subfigure}[b]{0.32\textwidth}
    \centering
    \includegraphics[width=\textwidth]{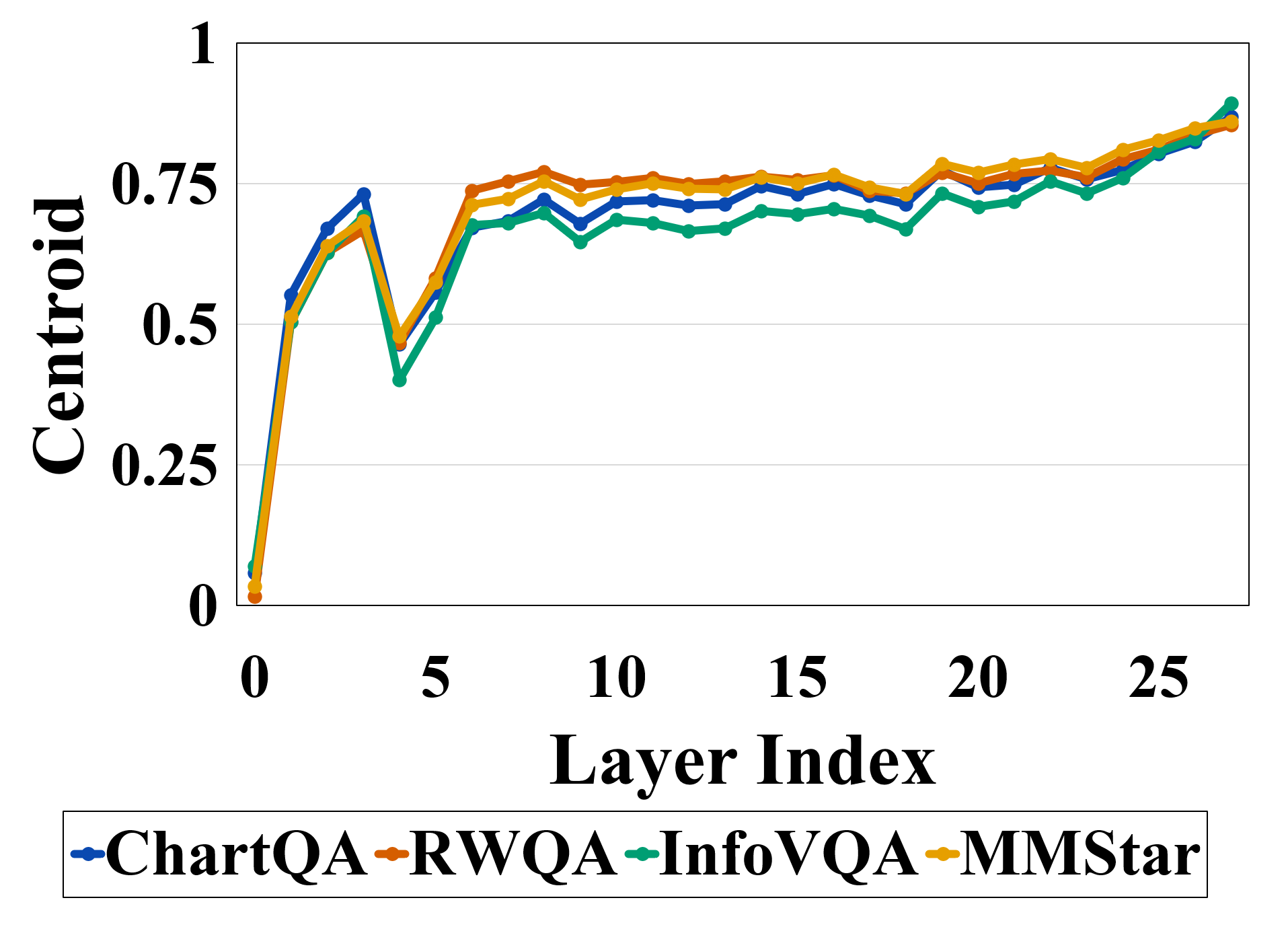}
    \caption{
    Vision–text centroid similarity
    }
    \label{fig:stage2-a}
  \end{subfigure}\hfill
  \begin{subfigure}[b]{0.32\textwidth}
    \centering
    \includegraphics[width=\textwidth]{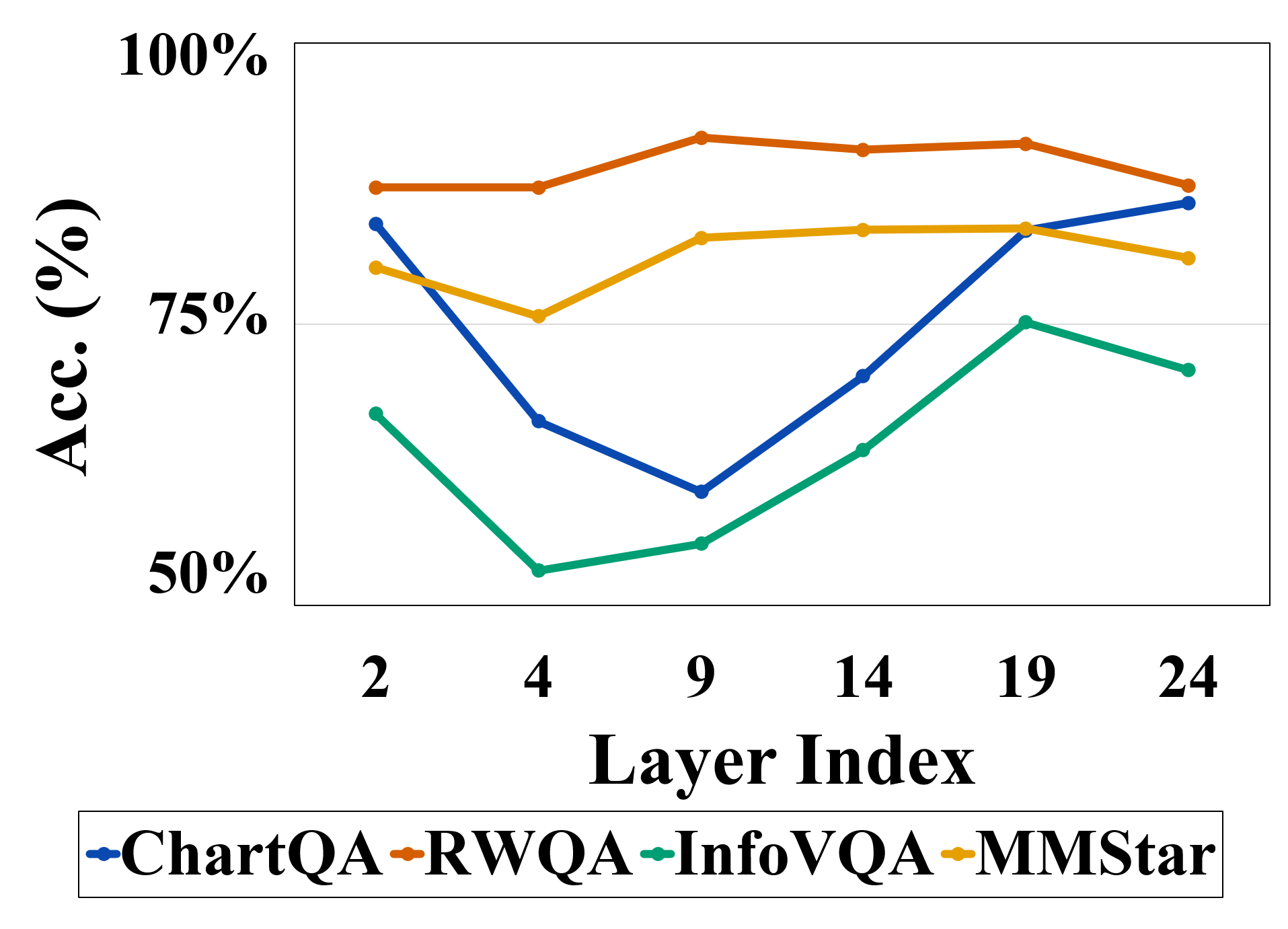}
    \caption{Deep-layer selection}
    \label{fig:stage2-b}
  \end{subfigure}\hfill
  \begin{subfigure}[b]{0.32\textwidth}
    \centering
    \includegraphics[width=\textwidth]{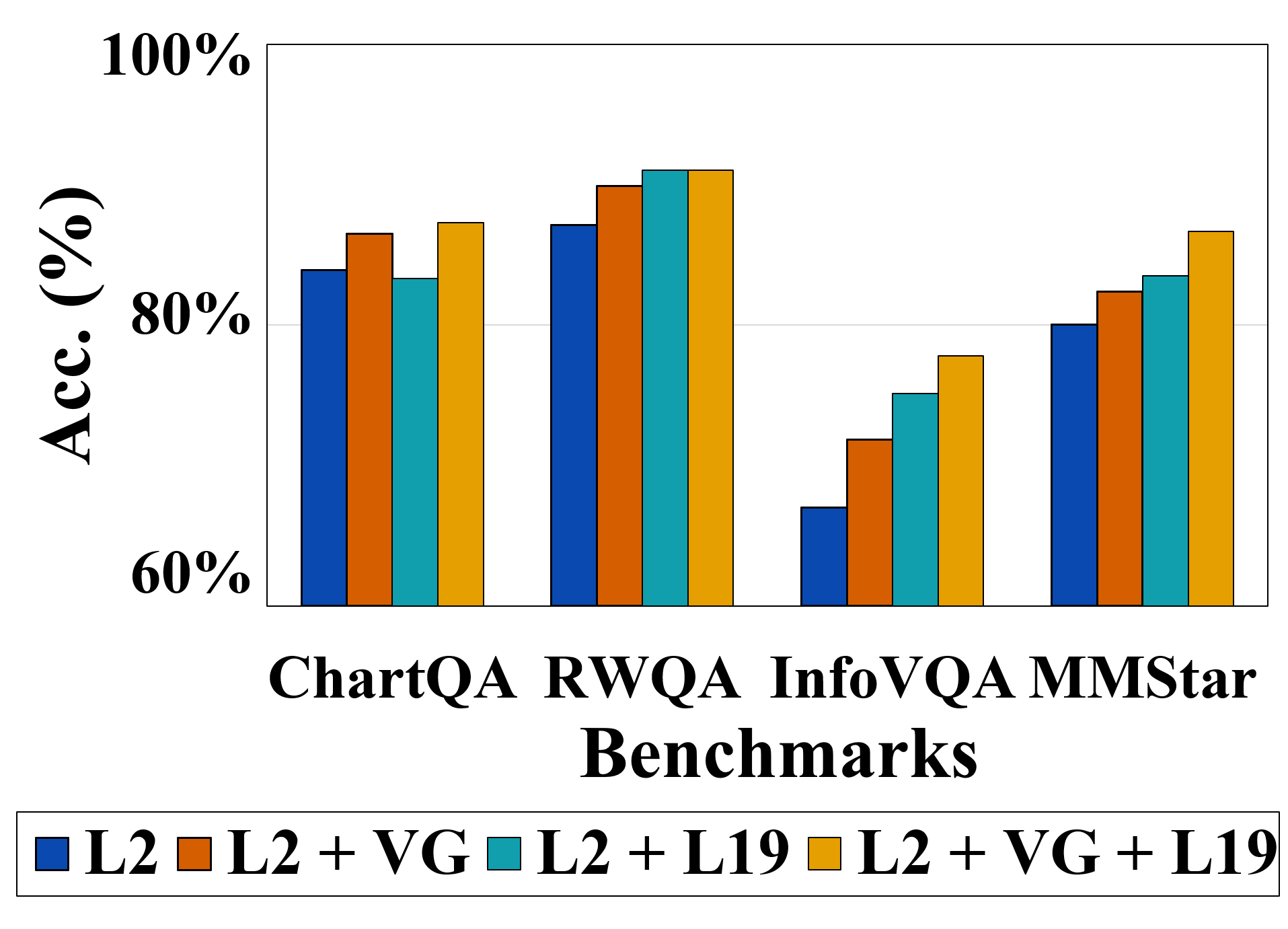}
    \caption{Stage 2 ablation}
    \label{fig:stage2-c}
  \end{subfigure}
  \caption{Layer-wise analysis on Qwen2.5-VL-7B:
  (a) vision–text centroid similarity,
  (b) retained accuracy across deep pruning layers under a 25\% layer-averaged token budget, 
  and (c) Stage 2 component analysis, where L2 denotes shallow instruction-aware pruning, VG adds Stage 1 visual guidance, and L19 denotes additional deep pruning at layer 19. Stage 1 pruning is disabled in (b)/(c).
  }
  \label{fig:stage2}
\end{figure}
To characterize the layer-wise interaction between the two modalities, we use a centroid-based proxy for vision–text alignment, defined as the cosine similarity between their mean representations at each decoder-layer input.
Let $X_t^{(\ell)}\in\mathbb{R}^{n_t\times d}$ and $X_v^{(\ell)}\in\mathbb{R}^{n_v\times d}$ denote the text and vision representations entering layer $\ell$, respectively. Their centroids and vision–text centroid similarity are defined as
\begin{equation}
\label{eq:align_score}
\bar{x}_t^{(\ell)}=\frac{1}{n_t}\sum_{i=1}^{n_t} X_{t,i}^{(\ell)},
\qquad
\bar{x}_v^{(\ell)}=\frac{1}{n_v}\sum_{j=1}^{n_v} X_{v,j}^{(\ell)},
\qquad
S^{(\ell)}=\frac{\bar{x}_t^{(\ell)} \cdot \bar{x}_v^{(\ell)}}{\|\bar{x}_t^{(\ell)}\|_2\|\bar{x}_v^{(\ell)}\|_2}.
\end{equation}
Throughout this section, we use samples correctly answered by the base model and define retained accuracy as the fraction remaining correct after compression. Fig.~\ref{fig:stage2-a} shows an early rise, subsequent drop, and eventual recovery in this proxy, broadly 
consistent with the hierarchical processing described above. This trend also appears across model sizes and architectures (Appendix~\ref{sec:appx-modality}). 
This pattern motivates examining an early pruning point before the drop and a later point after recovery.

\paragraph{Hierarchical context-aware pruning} \label{sec:stage2-dual}
Guided by this observation, 
we evaluate shallow and deep pruning points.
We fix the shallow pruning point at layer 2, which lies before the observed  
similarity drop, and sweep the deep pruning point under a 25\% layer-averaged budget to evaluate candidate deep pruning layers (Fig.~\ref{fig:stage2-b}).
Accuracy decreases when the second pruning point lies in the post-drop region and generally recovers as it moves deeper. However, pruning too late can reduce accuracy because meeting the same budget requires more aggressive late-stage pruning.
Given these pruning points, ERASE ranks vision tokens by instruction relevance at each stage.
For compatibility with FlashAttention-2, we separately compute the instruction-to-vision attention. The relevance of vision token $v$ is
\begin{equation}
\label{eq:stage2_attn}
a_v=\frac{1}{N_h}\sum_{h=1}^{N_h}\sum_{t \in \mathcal{T}_{\mathrm{inst}}}A_{t,v}^{(h)},
\end{equation}
where $N_h$ is the number of attention heads and $\mathcal{T}_{\mathrm{inst}}$ denotes the instruction tokens.
At the shallow pruning stage, ERASE combines instruction relevance with the visual cues from Stage 1. Since this is the first stage where instruction relevance is introduced, retaining visual informativeness helps avoid overly aggressive instruction-driven pruning at an early layer.
Specifically, shallow pruning reuses the Stage 1 edge density
$\rho_v=\frac{1}{p_h p_w}\sum_{(x,y)\in p(v)}e(x,y)$
and patch entropy $H_{p(v)}$, where $p(v)$ denotes the
corresponding Stage 1 patch.
The visual cues are combined with attention as
\begin{equation}
\label{eq:stage2_score}
g_v=\frac{\rho_v}{\mathrm{std}(\rho)}+\frac{H_{p(v)}}{\mathrm{std}(H)},
\qquad
z_v=a_v+w\,\frac{\mathrm{std}(a)}{\mathrm{std}(g)}g_v.
\end{equation}
To account for per-image scale differences, we normalize each visual cue by its standard deviation, which also makes the result independent of the units of $\rho$ and $H$, and rescale their combined guidance to match the spread of attention scores, allowing $w$ to control its relative strength.
Tokens are ranked by $z_v$ at the shallow layer and $a_v$ at the deep layer.
After shallow pruning has jointly considered visual informativeness and instruction relevance, deep pruning focuses on instruction relevance alone.
Fig.~\ref{fig:stage2-c} analyzes the components of Stage 2. Adding visual guidance to shallow pruning improves average accuracy by 3.14 points across four datasets. 
Hierarchical shallow–deep pruning further improves accuracy under the same layer-averaged budget.
Combining visually guided shallow pruning with deep instruction-aware pruning achieves the highest overall accuracy, supporting the proposed hierarchical design.

\section{Experiments}
\label{sec:exp}
\subsection{Experiment settings}\label{sec:setup}
\textbf{Models and benchmarks}
ERASE is evaluated on Qwen2.5-VL-7B~\citep{Qwen2_5_VL} and Qwen3-VL-8B~\citep{Qwen3-VL} with native-resolution image processing, and InternVL3-8B~\citep{zhu2025internvl3} with dynamic image tiling.
We evaluate using VLMEvalKit~\citep{duan2024vlmevalkit}: text recognition (TextVQA~\citep{textvqa}), document/chart understanding (ChartQA~\citep{masry-etal-2022-chartqa}, DocVQA~\citep{docvqa}, InfoVQA~\citep{infovqa}), visual reasoning (MathVista~\citep{lu2024mathvista}), and general visual question answering (RealWorldQA~\citep{xai2024grok15v_rwqa}, MMStar~\citep{chen2024we_mmstar}, MMBench~v1.1~\citep{liu2024mmbench}). 
Additional results on smaller models are provided in Appendix~\ref{sec:appx-exp_small}.

\textbf{Implementation details}
ERASE prunes at layer 2 and at approximately two-thirds of the decoder depth, using the same relative-depth rule across architectures: layer 19 of 28 for Qwen2.5-VL-7B and InternVL3-8B, and layer 24 of 36 for Qwen3-VL-8B. We set $\tau=0.45$ and $w=0.2$ for all models.
We compare against PDrop~\citep{Xing_2025_CVPR_Pdrop}, FastV~\citep{fastv}, DART~\citep{wen2025stop}, CDPruner~\citep{zhang2025cdpruner}, PruneSID~\citep{fang2026prunesid}, iLLaVA~\citep{hu2026illava}, AgilePruner~\citep{baek2026agilepruner}, VScan~\citep{zhang2026vscan}, and Nüwa~\citep{huang2026nwa} on InternVL3-8B. 
AgilePruner, VScan, and Nüwa are evaluated only on InternVL3-8B because they incur OOM on the Qwen models due to encoder attention maps.
Retain denotes the layer-averaged fraction of visual tokens processed during prefill relative to the base model, while Rel. denotes the average accuracy relative to the base model.
PDrop follows its reported configuration because its pruning schedule cannot match our retention ratios.
Appendix~\ref{sec:appx-setup} provides further details.


\begin{table*}[!b]
\centering
\caption{Results on Qwen2.5-VL-7B.}
\label{tab:qwen2_5_7b}
\resizebox{\textwidth}{!}{
\begin{tabular}{l l|c c c c c c c c|c c}
\toprule
\textbf{Retain$\downarrow$}
& \textbf{Method}
& \textbf{TextVQA}
& \textbf{ChartQA}
& \textbf{DocVQA}
& \textbf{InfoVQA}
& \textbf{MathVista}
& \textbf{MMB\textsubscript{v1.1}}
& \textbf{RWQA}
& \textbf{MMStar}
& \textbf{Avg.$\uparrow$}
& \textbf{Rel. (\%)$\uparrow$} \\
\midrule

100\% & Base
& 85.30 & 86.00 & 94.82 & 82.31 & 65.10 & 82.51 & 70.20 & 65.53 & 78.97 & 100\\

\midrule

47\% & PDrop
& 84.19 & 82.12 & 92.55 & 71.34 & 56.40 & 77.48 & 64.97 & 58.87 & 73.49 & 93.06 \\

\midrule

\multirow{6}{*}{45\%}
& FastV
& 84.40 & 82.92 & 93.19 & 74.90 & 63.20 & 81.58 & 68.50 & 62.00 & 76.34 & 96.66 \\
& DART
& 80.38& 77.04& 85.35& 60.57& 62.00& 80.96& 69.15& 62.53& 72.25& 91.49\\
& CDPruner
& 80.96 & 66.04 & 85.41 & 61.91 & 56.70 & 80.88 & 67.84 & 63.07 & 70.35 & 89.08 \\
& PruneSID
& 80.46 & 80.56 & 90.32 & 64.95 & 61.30 & 81.11 & 69.15 & 62.73 & 73.82 & 93.48 \\
& iLLaVA
& 84.16 & 82.28 & 93.09 & 76.43 & 64.30 & 81.81 & 68.76 & 62.27 & \underline{76.64}& \underline{97.04}\\
& ERASE
& 84.52 & 83.24 & 94.22 & 79.27 & 64.20 & 81.27 & 70.20 & 64.13 & \textbf{77.63} & \textbf{98.30} \\
\midrule
\multirow{6}{*}{35\%}
& FastV
& 83.75 & 79.04 & 90.27 & 68.13 & 62.30 & 80.80 & 67.97 & 60.20 & 74.06 & 93.78 \\
& DART
& 76.91& 70.40& 78.17& 53.10& 57.70& 79.88& 67.19& 61.33& 68.09& 86.21\\
& CDPruner
& 79.19 & 59.24 & 81.57 & 57.56 & 55.40 & 80.42 & 67.19 & 62.00 & 67.82 & 85.88 \\
& PruneSID
& 77.04 & 77.60 & 86.07 & 57.49 & 60.40 & 80.50 & 67.84 & 62.20 & 71.14 & 90.09 \\
& iLLaVA
& 83.24 & 79.08 & 90.91 & 72.11 & 60.70 & 81.81 & 69.02 & 62.40 & \underline{74.91}& \underline{94.86}\\
& ERASE
& 84.26 & 83.04 & 94.12 & 76.89 & 63.70 & 81.66 & 69.80 & 63.93 & \textbf{77.18} & \textbf{97.73} \\
\midrule
\multirow{6}{*}{25\%}
& FastV
& 82.01 & 71.24 & 81.09 & 55.84 & 58.20 & 78.72 & 66.41 & 56.73 & 68.78 & 87.09 \\
& DART
& 69.56& 59.16& 66.25& 41.83& 54.00& 78.56& 64.44& 57.07& 61.36& 77.70\\
& CDPruner
& 75.92 & 50.52 & 75.65 & 51.09 & 53.30 & 79.80 & 65.49 & 60.53 & 64.04 & 81.09 \\
& PruneSID
& 70.65 & 70.84 & 77.59 & 50.13 & 59.20 & 79.49 & 65.62 & 60.80 & 66.79 & 84.58 \\
& iLLaVA
& 80.64 & 70.76 & 86.26 & 64.74 & 58.80 & 81.66 & 68.50 & 61.60 & \underline{71.62}& \underline{90.69}\\
& ERASE
& 83.50 & 80.20 & 93.35 & 71.83 & 62.20 & 81.50 & 69.41 & 62.60 & \textbf{75.57} & \textbf{95.70} \\
\bottomrule
\end{tabular}
}
\end{table*}

\begin{table*}[t]
\centering
\caption{Results on Qwen3-VL-8B.}
\label{tab:qwen3_8b}
\resizebox{\textwidth}{!}{
\begin{tabular}{l l|c c c c c c c c|c c}
\toprule
\textbf{Retain$\downarrow$}
& \textbf{Method}
& \textbf{TextVQA}
& \textbf{ChartQA}
& \textbf{DocVQA}
& \textbf{InfoVQA}
& \textbf{MathVista}
& \textbf{MMB\textsubscript{v1.1}}
& \textbf{RWQA}
& \textbf{MMStar}
& \textbf{Avg.$\uparrow$}
& \textbf{Rel. (\%)$\uparrow$} \\
\midrule

100\% & Base
& 82.98 & 83.16 & 95.75 & 83.21 & 69.10 & 84.52 & 71.37 & 66.60 & 79.59 & 100\\
\midrule
47\% & PDrop
& 75.05 & 61.64 & 84.94 & 59.46 & 48.80 & 77.94 & 68.76 & 54.93 & 66.44 & 83.48 \\
\midrule
\multirow{6}{*}{45\%}
& FastV
& 75.18 & 53.96 & 81.92 & 55.22 & 49.00 & 76.63 & 68.24 & 55.07 & 64.40 & 80.92 \\
& DART
& 71.53& 69.32& 82.69& 59.65& 56.10& 79.41& 67.84& 59.53& 68.26& 85.77\\
& CDPruner
& 74.24 & 56.00 & 72.63 & 52.28 & 42.60 & 77.55 & 63.40 & 52.93 & 61.45 & 77.22 \\
& PruneSID
& 78.18 & 76.92 & 93.46 & 73.16 & 59.90 & 78.87 & 69.15 & 60.40 & \underline{73.76}& \underline{92.67}\\
& iLLaVA
& 77.40 & 65.64 & 93.02 & 76.57 & 45.40 & 81.35 & 70.85 & 60.53 & 71.35 & 89.64 \\
& ERASE
& 80.99 & 77.76 & 94.77 & 77.74 & 54.70 & 81.19& 70.33 & 61.53 & \textbf{74.88}& \textbf{94.08}\\
\midrule
\multirow{6}{*}{35\%}
& FastV
& 69.77 & 42.00 & 70.07 & 45.61 & 43.60 & 74.61 & 65.75 & 49.00 & 57.55 & 72.31 \\
& DART
& 65.18& 60.36& 73.30& 50.36& 49.80& 78.25& 68.76& 55.07& 62.64& 78.70\\
& CDPruner
& 69.94 & 46.72 & 63.57 & 47.70 & 38.00 & 73.14 & 61.44 & 49.33 & 56.23 & 70.65 \\
& PruneSID
& 73.54 & 65.64 & 90.46 & 66.90 & 52.80 & 75.62 & 68.24 & 60.33 & \underline{69.19}& \underline{86.94}\\
& iLLaVA
& 74.18 & 62.28 & 90.74 & 70.59 & 44.30 & 80.19 & 70.33 & 57.93 & 68.82 & 86.47 \\
& ERASE
& 79.99 & 75.04 & 94.29 & 72.82 & 51.00 & 79.49& 70.85 & 59.07 & \textbf{72.82}& \textbf{91.50}\\
\midrule
\multirow{6}{*}{25\%}
& FastV
& 61.48 & 27.76 & 51.42 & 35.66 & 37.80 & 68.96 & 63.40 & 45.80 & 49.04 & 61.61 \\
& DART
& 57.26& 50.20& 57.46& 40.48& 47.60& 74.46& 64.71& 51.87& 55.51& 69.74\\
& CDPruner
& 62.26 & 37.92 & 52.07 & 41.39 & 38.40 & 68.96 & 58.43 & 46.60 & 50.75 & 63.77 \\
& PruneSID
& 69.01 & 58.12 & 83.66 & 57.26 & 50.40 & 72.76 & 68.37 & 55.07 & \underline{64.33}& \underline{80.83}\\
& iLLaVA
& 70.61 & 51.04 & 85.81 & 58.54 & 42.50 & 78.02 & 67.58 & 55.13 & 63.65 & 79.98 \\
& ERASE
& 77.75 & 68.08 & 92.50 & 65.13 & 47.10 & 76.63& 69.28 & 55.60 & \textbf{69.01}& \textbf{86.71}\\

\bottomrule
\end{tabular}
}
\end{table*}

\begin{table*}[t]
\centering
\caption{Results on InternVL3-8B.}
\label{tab:internvl3_8b}
\resizebox{\textwidth}{!}{
\begin{tabular}{l l|c c c c c c c c|c c}
\toprule
\textbf{Retain$\downarrow$}
& \textbf{Method}
& \textbf{TextVQA}
& \textbf{ChartQA}
& \textbf{DocVQA}
& \textbf{InfoVQA}
& \textbf{MathVista}
& \textbf{MMB\textsubscript{v1.1}}
& \textbf{RWQA}
& \textbf{MMStar}
& \textbf{Avg.$\uparrow$}
& \textbf{Rel. (\%)$\uparrow$} \\
\midrule

100\% & Base
& 82.11 & 86.08 & 91.98 & 75.49 & 68.80 & 85.76 & 71.24 & 68.73 & 78.77 & 100\\
\midrule
47\% & PDrop
& 81.46& 82.56& 88.69& 67.52& 58.50& 83.90& 70.59& 65.27& 74.81& 94.97\\
\midrule
\multirow{9}{*}{45\%}
& FastV
& 81.05& 82.52& 82.17& 64.37& 64.00& 84.98& 71.24& 66.00& 74.54& 94.63\\
& DART
& 74.19 & 78.72 & 77.59 & 57.33 & 66.30 & 83.90 & 70.33 & 66.00 & 71.80 & 91.14 \\
& CDPruner
& 70.78 & 64.80 & 66.87 & 53.99 & 57.80 & 82.04 & 66.27 & 60.80 & 65.42 & 83.05 \\
& PruneSID
& 72.96 & 77.76 & 76.02 & 58.93 & 66.10 & 82.82 & 70.59 & 64.93 & 71.26 & 90.47 \\
& iLLaVA
& 66.63 & 59.64 & 68.65 & 54.53 & 58.40 & 81.81 & 68.37 & 62.60 & 65.08 & 82.61 \\
& AgilePruner
& 79.33 & 79.60 & 83.83 & 66.82 & 63.40 & 83.90 & 69.15 & 64.73 & 73.85 & 93.74 \\
& VScan
& 80.89 & 81.04 & 88.73 & 71.44 & 60.90 & 83.98 & 71.24 & 65.40 & 75.45 & 95.78 \\
& Nüwa
& 80.54 & 82.96 & 87.13 & 70.34 & 65.90 & 84.06 & 71.37 & 66.27 & \underline{76.07}& \underline{96.57}\\
& ERASE
& 80.47 & 82.64 & 87.95 & 70.24& 66.40 & 84.75& 71.37 & 67.13 & \textbf{76.37}& \textbf{96.95}\\
\midrule
\multirow{9}{*}{35\%}
& FastV
& 79.88 & 79.84 & 73.80 & 54.69 & 64.30& 84.83& 70.33& 65.47& 71.64& 90.95\\
& DART
& 70.55 & 73.28 & 68.95 & 48.99& 65.10 & 83.36 & 68.63 & 63.60 & 67.81 & 86.08 \\
& CDPruner
& 65.28 & 55.96 & 55.42 & 46.92 & 52.10 & 80.57 & 63.27 & 57.60 & 59.64 & 75.71 \\
& PruneSID
& 67.42 & 72.28 & 62.56 & 49.84 & 65.10 & 81.97 & 68.89 & 64.33 & 66.55 & 84.48 \\
& iLLaVA
& 63.61 & 56.08 & 62.77 & 48.68 & 55.90 & 79.72 & 66.53 & 61.13 & 61.80 & 78.46 \\
& AgilePruner
& 76.71 & 76.72 & 77.61 & 61.33 & 60.10 & 83.44 & 67.19 & 62.13 & 70.65 & 89.69 \\
& VScan
& 79.71 & 79.28 & 85.06 & 68.28 & 57.30 & 83.13 & 70.07 & 64.07 & 73.36 & 93.13 \\
& Nüwa
& 79.33 & 79.84 & 82.57 & 66.77 & 64.00 & 83.51 & 70.46 & 65.20 & \underline{73.96}& \underline{93.89}\\
& ERASE
& 79.92& 80.72 & 86.53 & 65.30& 65.80 & 84.21& 71.37 & 66.73 & \textbf{75.07}& \textbf{95.30}\\
\midrule
\multirow{9}{*}{25\%}
& FastV
& 76.68 & 74.92 & 63.86 & 45.21 & 61.50 & 83.82 & 69.67 & 62.93 & 67.32 & 85.46 \\
& DART
& 64.08 & 62.08 & 54.94 & 40.18 & 62.40 & 80.88 & 66.41 & 60.33 & 61.41 & 77.96 \\
& CDPruner
& 58.21 & 46.76 & 42.60 & 39.46 & 42.50 & 78.25 & 60.39 & 55.93 & 53.01 & 67.30 \\
& PruneSID
& 57.10 & 61.88 & 46.75 & 40.19 & 58.20 & 78.87 & 67.84 & 61.87 & 59.09 & 75.01 \\
& iLLaVA
& 58.59 & 50.92 & 56.64 & 44.21 & 53.40 & 79.64 & 66.27 & 60.00 & 58.71 & 74.53 \\
& AgilePruner
& 71.13 & 67.28 & 66.77 & 52.76 & 57.60 & 83.44 & 64.58 & 59.40 & 65.37 & 82.98 \\
& VScan
& 76.35 & 72.60 & 75.02 & 59.64 & 53.60 & 81.58 & 66.54 & 61.73 & 68.38 & 86.81 \\
& Nüwa
& 75.94 & 74.92 & 73.19 & 59.76 & 57.90 & 82.66 & 68.76 & 62.93 & \underline{69.51}& \underline{88.24}\\
& ERASE
& 78.23 & 78.12 & 80.66 & 57.22& 63.30 & 84.37 & 70.98 & 66.20 & \textbf{72.39}& \textbf{91.89}\\
\bottomrule
\end{tabular}
}
\end{table*}

\subsection{Accuracy evaluation} \label{sec:acc}
ERASE achieves the highest average accuracy at all three matched retention ratios on Qwen2.5-VL-7B and Qwen3-VL-8B (Tables~\ref{tab:qwen2_5_7b} and~\ref{tab:qwen3_8b}). At 25\% retention, ERASE outperforms the strongest baseline on each model, iLLaVA and PruneSID, by 3.95 and 4.68 points, respectively, while preserving 95.70\% and 86.71\% of the base model accuracy. The gains are particularly pronounced on OCR-intensive tasks, where removing even a small text region can alter the recognized content. Table~\ref{tab:internvl3_8b} evaluates ERASE on InternVL3-8B, which uses dynamic tiling and thus tests its applicability to a different visual input architecture. Its encoder \texttt{[CLS]} token also enables comparison with AgilePruner, VScan, and Nüwa without OOM. VScan and Nüwa are the strongest baselines at matched retention ratios, yet ERASE achieves the highest average accuracy at all three ratios. At 25\% retention, ERASE preserves 91.89\% of the base-model accuracy and exceeds AgilePruner, VScan, and Nüwa by 7.02, 4.01, and 2.88 points in average accuracy, respectively. Appendix~\ref{sec:appx-rescale} additionally evaluates AgilePruner, VScan, and Nüwa on a Qwen model using downscaled images.

\subsection{Ablation and efficiency analysis} \label{sec:ablation}
\begin{figure*}
  \centering
  \begin{subfigure}[t]{0.3\textwidth}
  \centering
    \includegraphics[width=\linewidth]{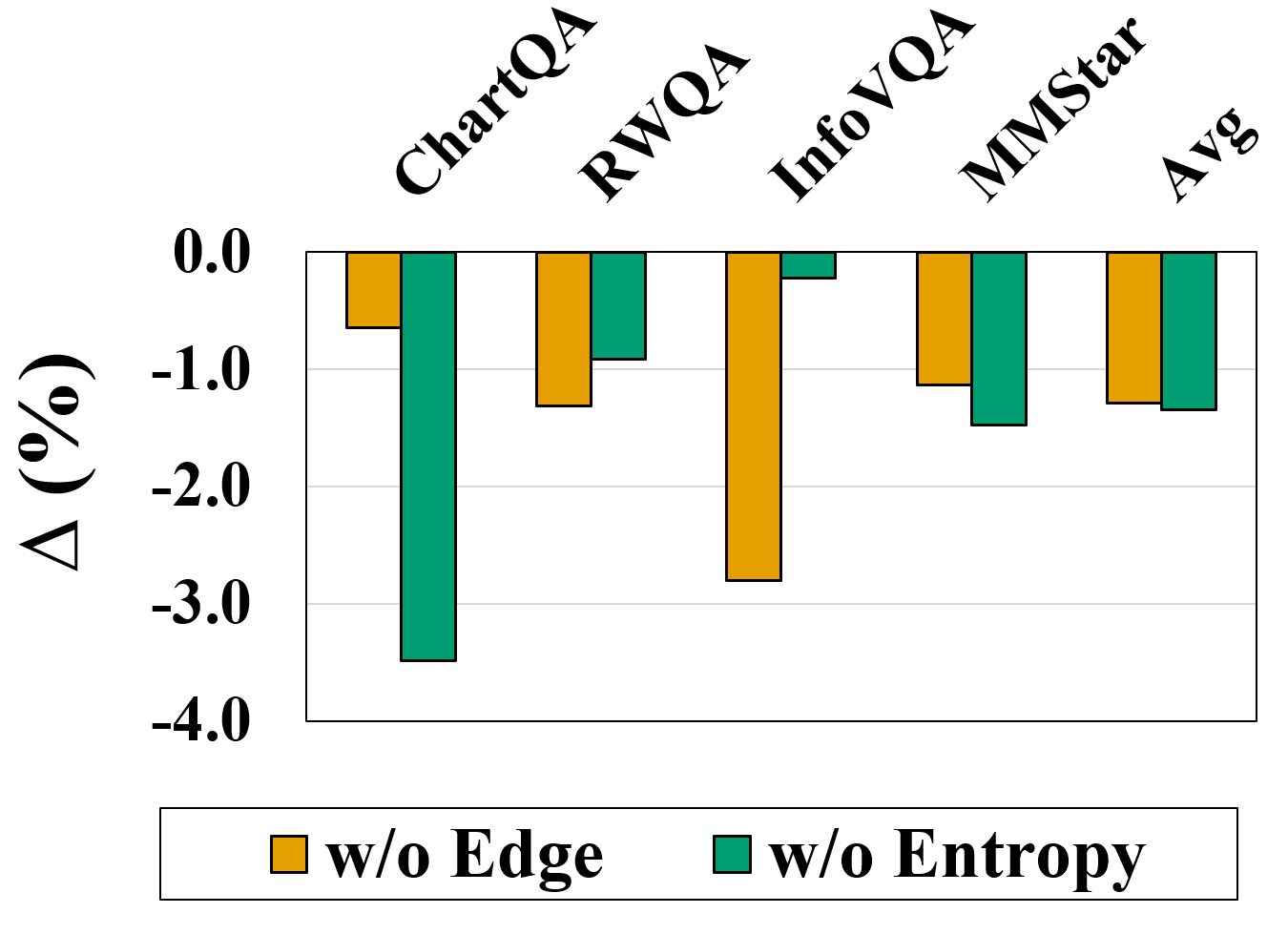}
    \caption{Stage 1 visual cues
    }
    \label{fig:ablation-a}
  \end{subfigure}
  \hfill
  \begin{subfigure}[t]{0.3\textwidth}
    \includegraphics[width=\linewidth]{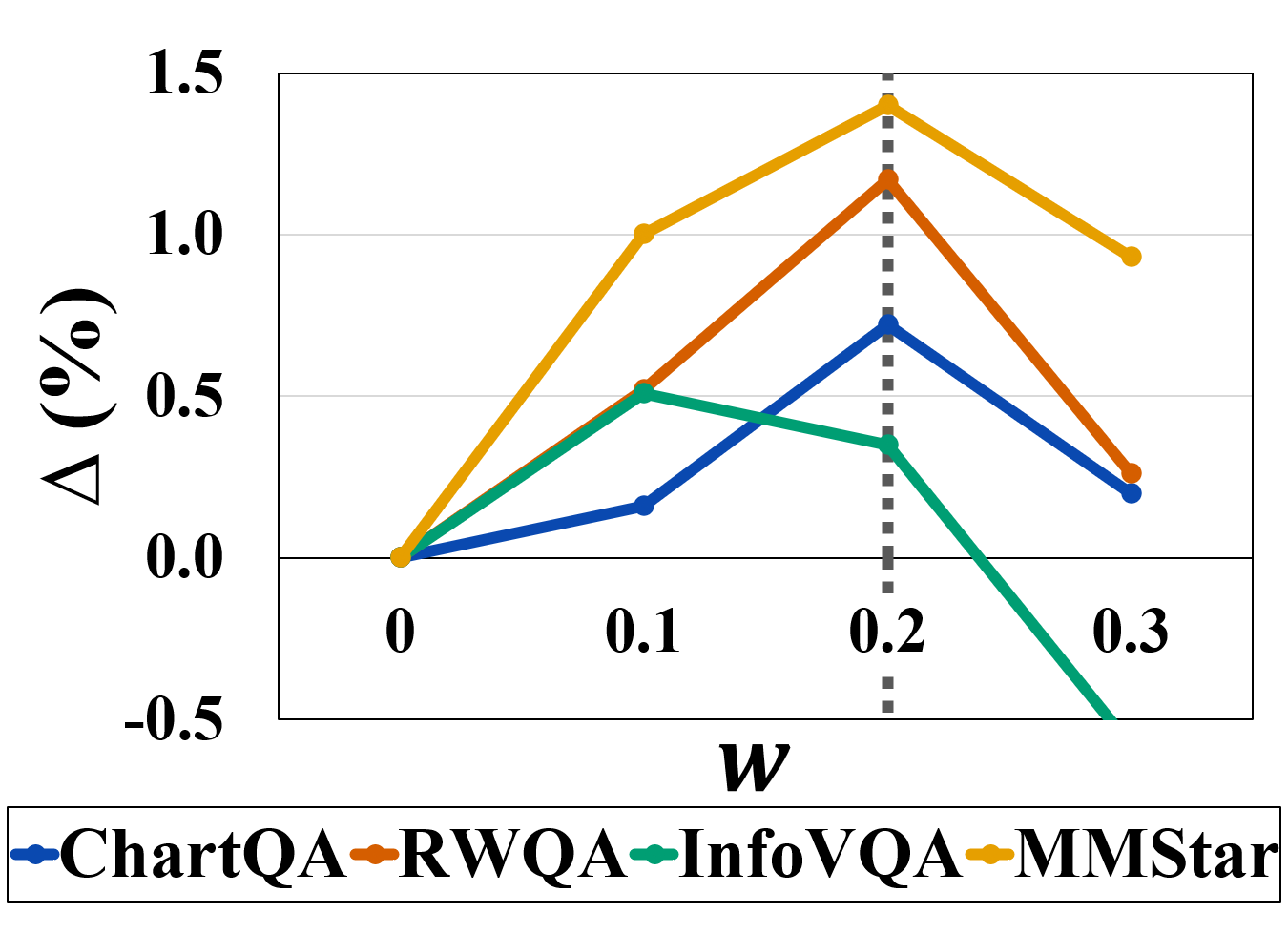}
    \caption{Stage 2 guidance weight
    }
    \label{fig:ablation-b}
  \end{subfigure}
  \hfill
  \begin{subfigure}[t]{0.38\textwidth}
    \includegraphics[width=\linewidth]{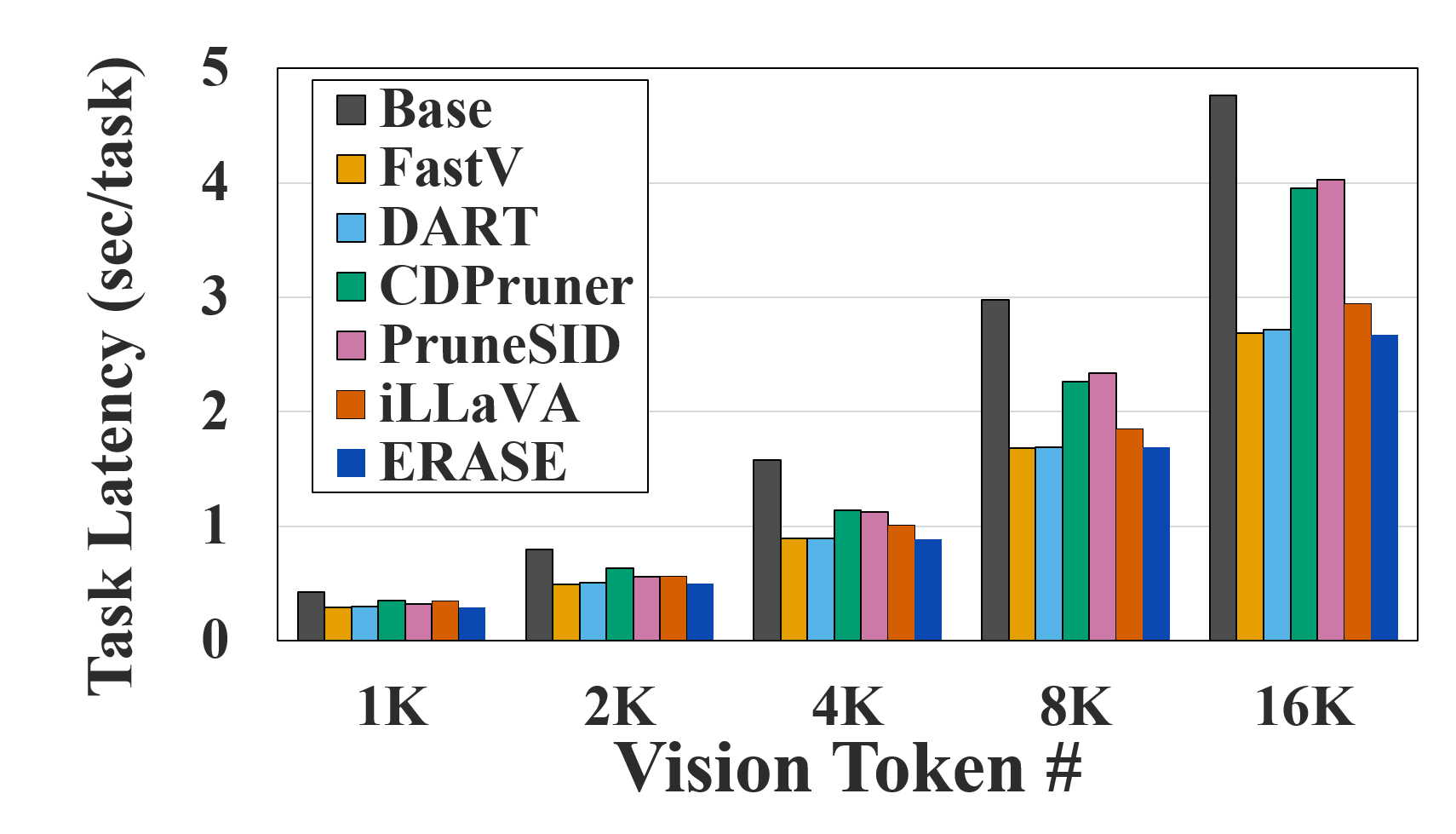}
    \caption{Task latency}
    \label{fig:tasktime}
  \end{subfigure}
  \vspace{-6pt}
  \caption{
    Ablation and efficiency analysis on Qwen2.5-VL-7B.
    (a) Stage 1 accuracy loss using either visual cue alone.
    (b) Stage 2 shallow-layer visual-guidance accuracy gains over $w=0$. Accuracies at $w=0$ are 79.48, 68.24, 71.48, and 61.20 for ChartQA, RWQA, InfoVQA, and MMStar, respectively.
    (c) End-to-end latency across different visual token counts.
  }
  \label{fig:ablation}
\end{figure*}

\paragraph{Effectiveness of Stage 1 pruning}
We compare ERASE's Stage 1 with vision-only compression methods, PruneSID and Stage 1 of iLLaVA at the benchmark-wise average retention ratios obtained by ERASE's adaptive Stage 1 (Table~\ref{tab:stage1-compare}). Fixed ERASE matches these ratios by removing or adding patches to its Stage 1 selection using the visual score in Eq.~\ref{eq:stage2_score}. It achieves an average accuracy of 77.70\%, outperforming iLLaVA (76.89\%) and PruneSID (76.06\%). Adapting retention to individual images using image-relative statistics further improves average accuracy to 78.22\%. These results support the effectiveness of ERASE’s visual selection criteria and the additional benefit of image-adaptive retention.
\begin{table*}[t]
\centering
\caption{Vision-only token compression comparison. $^{\dagger}$ denotes Stage 1 only.
}
\label{tab:stage1-compare}
\resizebox{\linewidth}{!}{
\begin{tabular}{l c|c c c c c c c c|c c}
\toprule
\textbf{Method} & \textbf{Budget}
& \textbf{TextVQA} & \textbf{ChartQA} & \textbf{DocVQA} & \textbf{InfoVQA}
& \textbf{MathVista} & \textbf{MMB\textsubscript{v1.1}} & \textbf{RWQA} & \textbf{MMStar}
& \textbf{Avg.$\uparrow$} & \textbf{Rel. (\%)$\uparrow$} \\
\midrule
Base & --
& 85.30 & 86.00 & 94.82 & 82.31 & 65.10 & 82.51 & 70.20 & 65.53 & 78.97 & 100 \\
\midrule
PruneSID & Fixed
& 84.38 & 82.48 & 89.28 & 73.50 & 62.90 & 81.66 & 70.20 & 64.07 & 76.06 & 96.31 \\
iLLaVA$^{\dagger}$ & Fixed
& 84.66 & 83.44 & 91.06 & 79.62 & 61.70 & 82.28 & 68.50 & 63.87 & 76.89 & 97.37 \\
ERASE$^{\dagger}$ & Fixed
& 84.81 & 83.56 & 93.58 & 79.31 & 64.40 & 82.59 & 70.07 & 63.27 & \underline{77.70} & \underline{98.39} \\
ERASE$^{\dagger}$ & Adaptive
& 84.89 & 83.76 & 94.32 & 80.31 & 65.00 & 81.73 & 71.11 & 64.60 & \textbf{78.22} & \textbf{99.04} \\
\midrule
\multicolumn{2}{l|}{Retain (\%)}
& 69.11 & 52.51 & 41.28 & 60.77 & 47.22 & 63.84 & 64.73 & 57.24 & 57.09 & -- \\
\bottomrule
\end{tabular}
}
\end{table*}
To assess the contribution of each cue at Stage 1, we compare the combined edge-entropy criterion with edge-only and entropy-only variants in Stage 1 (Fig.~\ref{fig:ablation-a}). Either cue alone yields lower accuracy than both cues together across all benchmarks, with their relative contributions varying by benchmark. These results support retaining regions identified by either cue.

\paragraph{Impact of visual guidance at Stage 2}
Fig.~\ref{fig:ablation-b} varies $w$ at the shallow pruning layer under 25\% retention with Stage 1 enabled. Moderate visual guidance generally improves accuracy, with the best average performance at $w=0.2$, whereas stronger guidance degrades performance across most benchmarks. This suggests that visual cues complement instruction relevance at the shallow layer by preserving useful visual context, while excessive visual guidance can dominate token selection and weaken instruction-aware pruning. We therefore set $w=0.2$ to balance visual guidance and instruction relevance.


\begin{table}[b]
  \caption{
  Efficiency comparison of pruning methods on 4K images with 16K vision tokens.
  }
  \label{tab:efficiency_metrics}
  \centering
    \resizebox{0.85\textwidth}{!}{ 
  \begin{tabular}{l| c c c c}
    \toprule
    \multirow{2}{*}{\textbf{Method}} & \textbf{KV Cache} & \textbf{Prefill Latency} & \textbf{Decode Latency}& \textbf{Task Latency} 
\\
    & (MB) $\downarrow$ & (ms) $\downarrow$ & (ms/token) $\downarrow$& (ms/task) $\downarrow$
\\
    \midrule
    Base & 890.74& 4647& 37.34& 4762\\
    \midrule
    FastV & 224.59 ($\times$3.97)& 2582 ($\times$1.80)& 33.47 ($\times$1.12)& 2689 ($\times$1.77)\\
    DART & 224.75 ($\times$3.96)& 2606 ($\times$1.78)& 35.24 ($\times$1.06)& 2715 ($\times$1.75)\\
    CDPruner & 224.61 ($\times$3.97)& 3848 ($\times$1.21)& 33.64 ($\times$1.11)& 3954 ($\times$1.20)\\
    PruneSID & 224.62 ($\times$3.97)& 3926 ($\times$1.18)& 31.78 ($\times$1.17)& 4029 ($\times$1.18)\\
 iLLaVA& 224.61 ($\times$3.97)& 2839 ($\times$1.64)& 33.54 ($\times$1.11)& 2941 ($\times$1.62)\\
    \textbf{ERASE (Ours)} & 224.60 ($\times$3.97)& 2569 ($\times$1.81)& 32.51 ($\times$1.15)& 2671 ($\times$1.78)\\
    \bottomrule
  \end{tabular}
    }
\end{table}

\paragraph{Inference efficiency} \label{sec:speed}
We profile Qwen2.5-VL-7B on a single NVIDIA RTX 5880 Ada Generation GPU with batch size 1 and FlashAttention-2~\citep{dao2023flashattention2}, using a randomly sampled 10\% subset of InfoVQA covering image resolutions corresponding to 1K--16K visual tokens. The target retention ratio is set to 25\%. As the number of visual tokens increases, ERASE achieves increasingly larger task-latency reductions over the base model (Fig.~\ref{fig:tasktime}). Similar retention ratios do not necessarily result in similar latency because pruning overhead varies across methods: CDPruner performs iterative subset selection in embedding space, 
while PruneSID relies on pairwise token similarity and non-maximum suppression, causing their overhead to grow with the visual token count. In contrast, ERASE and FastV use lightweight token-selection procedures and scale more efficiently. At 16K visual tokens (Table~\ref{tab:efficiency_metrics}), ERASE reduces the KV cache by $3.97\times$ and achieves $1.81\times$ prefill and $1.78\times$ end-to-end speedups with low decoding overhead. Its task latency remains comparable to the most efficient baselines while achieving substantially higher accuracy (Table~\ref{tab:qwen2_5_7b}). Appendix~\ref{sec:appx-efficiency} provides additional results averaged over different resolutions.

\section{Conclusion}

We proposed ERASE, an adaptive two-stage framework for efficient VLM inference. Stage 1 jointly determines visual token selection and image-dependent retention from lightweight raw-image statistics, while Stage 2 progressively removes instruction-irrelevant tokens through hierarchical instruction-aware pruning. Across diverse VLM architectures and benchmarks, ERASE achieves better efficiency--accuracy trade-offs than prior methods, preserving higher accuracy under aggressive compression while substantially reducing end-to-end latency for high-resolution inputs.


\subsection*{AI use statement}
In this work, we did not use generative AI tools for literature retrieval or discovery, drafting paper sections, developing theoretical models or conceptual frameworks, proposing or refining hypotheses, designing or providing feedback on research methodology or experiments, implementing methods, translation, or interpreting results. Generating synthetic datasets, formulating or proving mathematical claims, cleaning or reformatting datasets, and qualitative or thematic data analysis are not applicable to this work. Additionally, we used generative AI tools to polish author-written text for grammar, clarity, and readability. We reviewed all AI-assisted edits to ensure that they preserved the intended meaning and accurately reflected our methods and results. We take responsibility for the final content of this work, including text, claims, or artifacts produced with the aid of generative AI.

\subsection*{Reproducibility statement}
For reproducibility, we provide detailed descriptions of the experimental protocol and implementation settings used in this work. Section~\ref{sec:setup} and Appendix~\ref{sec:appx-setup} document the model and evaluation configurations, pruning-layer selection, retention settings, and hyperparameters required to reproduce our experiments. Algorithm~\ref{alg:erase} describes the complete end-to-end procedure of ERASE, covering both Stage 1 image-level pruning and Stage 2 hierarchical instruction-aware pruning. We additionally provide our source code in the supplementary material.

\bibliography{iclr2027_conference}
\bibliographystyle{iclr2027_conference}

\clearpage
\appendix
\section{Appendix}
\subsection{Detailed experimental settings}\label{sec:appx-setup}
\paragraph{Evaluation and baseline settings} 
We configure all baselines following their official implementations. 
CDPruner uses text tokens projected through the LLM embedding layer. DART and FastV prune at layer 2, with keep ratios adjusted to match the target layer-averaged retention. For iLLaVA, we use the official code setting for Qwen3-VL and the paper configuration for other model series. For Qwen3-VL, 40\%, 60\%, and 80\% of the pruning budget are allocated to Stage 1 at target retention ratios of 45\%, 35\%, and 25\%, respectively, with the remainder assigned to the LLM to achieve the target ratios. For VScan and Nüwa, Stage 2 is placed at the middle decoder layer. VScan uses $R_2=33.3\%$ with $R_1$ adjusted to the target, while Nüwa uses Stage 1 retention ratios of 55\%, 45\%, and 35\% for targets of 45\%, 35\%, and 25\%, respectively. PDrop cannot match the targets as its four-stage schedule begins pruning only after the first quarter of the decoder. We therefore use its original four-stage schedule, retaining 100\%, 50\%, 25\%, and 12.5\% of the vision tokens in successive stages. PruneSID and AgilePruner follow their official implementations, with pruning parameters adjusted to the target retention.
For benchmarks where VLMEvalKit recommends LLM-as-a-judge evaluation, we use Qwen1.5-1.8B-Chat~\citep{qwen_15}.

\paragraph{Configuration of ERASE}
For Stage 1, ERASE converts each input image to the luminance channel and applies $4\times4$ average pooling followed by a $3\times3$ Gaussian filter before computing the entropy and edge statistics. For Qwen-series models, which preserve the native image resolution, these statistics are computed directly from the input image. InternVL3 instead represents an image using a thumbnail together with multiple local tiles. Since the thumbnail and local tiles have different information densities, we compute their adaptive thresholds separately: the thumbnail uses its own mean entropy and median intensity, whereas all local tiles share the statistics computed over the tile set.

Because the Stage 1 retention ratio varies across images, ERASE dynamically selects one of three Stage 2 schedules to respect the target layer-averaged token budget. If Stage 1 already reaches or falls below the target retention, Stage 2 is skipped. 
In this case, the achieved retention can be lower than the target, so the reported retention ratios serve as upper bounds for ERASE. 
If using the default deep-layer keep ratio would require an early keep ratio greater than one, shallow pruning is skipped and the deep-layer keep ratio is adjusted instead. Otherwise, ERASE performs progressive pruning: the deep layer retains $30\%$ of the vision tokens reaching it, while the shallow-layer keep ratio is adjusted per sample to meet the target layer-averaged retention.

\subsection{Algorithm of ERASE}

\begin{algorithm}[t]
\caption{ERASE: Adaptive Two-Stage Vision Token Pruning}
\label{alg:erase}
\begin{algorithmic}[1]
\Require Image $I$, vision tokens $\mathbf{V}\in\mathbb{R}^{N\times d}$, instruction tokens $\mathbf{T}\in\mathbb{R}^{N_t\times d}$, target retention $\bar r$
\Require Pooling factor $f$, edge sensitivity $\tau$, number of decoder layers $L$, pruning layers $(l_e,l_l)$, default late keep ratio $r_l^{0}$, guidance weight $w$
\Ensure Pruned multimodal tokens $\mathbf{X}_{\mathrm{out}}$

\Statex \textbf{// Stage 1: Image-level pruning}
\State $\tilde{Y}\leftarrow
\mathrm{Gaussian}\!\left(\mathrm{AvgPool}_{f\times f}(\mathrm{Luminance}(I))\right)$
\State $H_p\leftarrow-\sum_{v}\hat P_p(v)\ln\hat P_p(v),\quad \bar H\leftarrow\frac{1}{N}\sum_{p=1}^{N}H_p$
\State $M\leftarrow\|\nabla_{\mathrm{Sobel}}\tilde Y\|_2,\quad
e(x,y)\leftarrow\mathbf{1}[M(x,y)>(1-\tau)\operatorname{median}(\tilde Y)]$
\State $\rho_v\leftarrow\frac{1}{p_hp_w}\sum_{(x,y)\in p(v)}e(x,y)$
\State $\mathcal{I}_1\leftarrow
\{v\mid H_{p(v)}>\bar H\ \lor\ \rho_v>0\},\quad
\mathbf{V}_1\leftarrow\mathbf{V}[\mathcal{I}_1,:],\quad
r_1\leftarrow|\mathcal{I}_1|/N$

\Statex \textbf{// Stage 2: Adaptive compression schedule}
\State $\displaystyle
\hat r_e\leftarrow
\frac{L\bar r/r_1-l_e}
{(l_l-l_e)+r_l^{0}(L-l_l)}$
\Comment{only if $r_1>\bar r$}
\State $\displaystyle
(r_e,r_l)\leftarrow
\begin{cases}
(1,1), & r_1\leq\bar r
\quad\text{(Stage 2 skipped)},\\[2pt]
\left(1,\frac{L\bar r/r_1-l_l}{L-l_l}\right),
& \hat r_e\geq1
\quad\text{(late only)},\\[4pt]
(\hat r_e,r_l^{0}), & \text{otherwise}
\quad\text{(progressive)}.
\end{cases}$

\Statex \textbf{// Hierarchical context-aware pruning}
\State $\mathbf{V}^{(0)}\leftarrow\mathbf{V}_1,\quad \mathbf{T}^{(0)}\leftarrow\mathbf{T}$
\For{$\ell=0,\ldots,L-1$}
    \State $[\mathbf{V}^{(\ell+1)};\mathbf{T}^{(\ell+1)}]
    \leftarrow\mathrm{Decoder}_{\ell}\big([\mathbf{V}^{(\ell)};\mathbf{T}^{(\ell)}]\big)$
    \If{$\ell+1\in\{l_e,l_l\}$}
        \State $\displaystyle
        a_v\leftarrow\frac{1}{N_h}
        \sum_{h=1}^{N_h}\sum_{t=1}^{N_t}A^{(h)}_{t,v}$
        \Comment{$N_v$: current number of vision tokens}
    \EndIf
    \If{$\ell+1=l_e$ and $r_e<1$}
        \State $\displaystyle
        g_v\leftarrow\frac{\rho_v}{\mathrm{std}(\rho)}
        +\frac{H_{p(v)}}{\mathrm{std}(H)},\quad
        z_v\leftarrow a_v+w\,\frac{\mathrm{std}(a)}{\mathrm{std}(g)}\,g_v,\quad
        \mathcal{I}_e\leftarrow\operatorname{TopK}(z,\lfloor r_eN_v\rfloor)$
        \State $\mathbf{V}^{(\ell+1)}\leftarrow\mathbf{V}^{(\ell+1)}[\mathcal{I}_e,:]$
    \ElsIf{$\ell+1=l_l$ and $r_l<1$}
        \State $\displaystyle
        \mathcal{I}_l\leftarrow\operatorname{TopK}(a,\lfloor r_lN_v\rfloor),\quad
        \mathbf{V}^{(\ell+1)}\leftarrow\mathbf{V}^{(\ell+1)}[\mathcal{I}_l,:]$
    \EndIf
\EndFor
\State \Return $\mathbf{X}_{\mathrm{out}}\leftarrow[\mathbf{V}^{(L)};\mathbf{T}^{(L)}]$
\end{algorithmic}
\end{algorithm}

Algorithm~\ref{alg:erase} summarizes the overall ERASE procedure. Stage 1 first converts the image to luminance and applies average pooling followed by Gaussian smoothing. It then identifies structurally informative regions using Sobel-based edges and locally informative regions using patch-wise entropy. A vision token is pruned only when its corresponding patch contains neither a detected edge nor above-average entropy.
After Stage 1, ERASE constructs a per-sample Stage 2 pruning schedule from the Stage 1 retention $r_1$ and target layer-averaged retention $\bar r$. Depending on $r_1$, Stage 2 is either skipped, applied only at the deep pruning point, or applied progressively at both shallow and deep pruning points. Shallow pruning combines instruction-to-vision attention with the visual cues obtained in Stage 1, whereas deep pruning relies on attention alone.

For instance, consider a 28-layer model with target layer-averaged retention $\bar r=25\%$ and layers 2 and 19 as the shallow and deep pruning points, respectively. If Stage 1 retains $30\%$ of the original vision tokens, the progressive solution would require $r_e>1$. ERASE therefore skips shallow pruning and uses only the deep pruning point, with $r_l\approx48\%$. Consequently, layers 0--18 operate with $30\%$ of the original vision tokens, while layers 19--27 operate with approximately $30\%\times48\%\approx14.4\%$. If Stage 1 retains $50\%$, ERASE performs progressive pruning with $r_e\approx61\%$ at layer 2 and the default $r_l^{0}=30\%$ at layer 19. Thus, layers 0--1 operate with $50\%$ of the original vision tokens, layers 2--18 with approximately $50\%\times61\%\approx30.5\%$, and layers 19--27 with approximately $50\%\times61\%\times30\%\approx9.2\%$. In both cases, the vision-token retention averaged across all 28 decoder
layers is approximately $25\%$.

\subsection{Extended evaluation results}\label{sec:appx-exp_small}
\begin{table*}[t]
\centering
\caption{Results on Qwen2.5-VL-3B.}
\label{tab:qwen2_5_3b}
\resizebox{\textwidth}{!}{
\begin{tabular}{l l|c c c c c c c c|c c}
\toprule
\textbf{Retain$\downarrow$}
& \textbf{Method}
& \textbf{TextVQA}
& \textbf{ChartQA}
& \textbf{DocVQA}
& \textbf{InfoVQA}
& \textbf{MathVista}
& \textbf{MMB\textsubscript{v1.1}}
& \textbf{RWQA}
& \textbf{MMStar}
& \textbf{Avg.$\uparrow$}
& \textbf{Rel. (\%)$\uparrow$} \\
\midrule

100\% & Base
& 79.06 & 83.96 & 93.07 & 75.95 & 60.30 & 77.48 & 66.41 & 55.93 & 74.02 & 100 \\

\midrule


47\% & PDrop
& 76.94 & 70.08 & 80.92 & 58.20 & 51.60 & 73.92 & 60.13 & 52.33 & 65.52 & 88.51 \\

\midrule

\multirow{6}{*}{45\%}
& FastV
& 77.53 & 80.08 & 89.86 & 67.50 & 58.30 & 76.47 & 66.27 & 53.00 & \underline{71.13}& \underline{96.09}\\

& DART
& 73.45& 79.80& 85.69& 53.96& 55.20& 75.23& 64.58& 54.20& 67.76& 91.55\\

& CDPruner
& 77.68 & 73.52 & 78.24 & 53.22 & 54.90 & 76.08 & 63.92 & 53.53 & 66.39 & 89.69 \\

& PruneSID
& 69.83 & 79.56 & 86.20 & 58.62 & 57.20 & 75.15 & 63.01 & 55.80 & 68.17 & 92.10 \\

& iLLaVA
& 77.09 & 75.64 & 89.09 & 67.39 & 58.10 & 76.32 & 65.10 & 55.13 & 70.48 & 95.22 \\

& ERASE
& 77.47 & 80.48 & 91.41 & 72.56 & 56.20 & 76.39 & 64.58& 55.20 & \textbf{71.79}& \textbf{96.98}\\

\midrule

\multirow{6}{*}{35\%}
& FastV
& 76.21 & 76.16 & 85.35 & 61.95 & 55.40 & 75.77 & 64.58 & 52.40 & \underline{68.48}& \underline{92.51}\\

& DART
& 70.12& 75.36& 78.54& 44.59& 52.80& 75.23& 61.70& 53.60& 63.99& 86.45\\

& CDPruner
& 77.21 & 69.40 & 79.81 & 52.18 & 51.20 & 75.46 & 62.48 & 51.40 & 64.89 & 87.67 \\

& PruneSID
& 63.44 & 75.60 & 81.54 & 51.96 & 56.30 & 73.30 & 64.31 & 54.67 & 65.14 & 88.00 \\

& iLLaVA
& 75.42 & 68.16 & 84.77 & 60.69 & 56.70 & 76.24 & 63.79 & 54.87 & 67.58 & 91.30 \\

& ERASE
& 76.52 & 78.76 & 90.99 & 70.22 & 55.60 & 75.93 & 64.84& 54.13 & \textbf{70.87}& \textbf{95.75}\\

\midrule

\multirow{6}{*}{25\%}
& FastV
& 72.93 & 65.52 & 73.39 & 50.74 & 51.60 & 72.68 & 61.70 & 50.80 & \underline{62.42}& \underline{84.33}\\

& DART
& 63.89& 67.36& 66.58& 34.86& 50.20& 72.99& 59.87& 50.67& 58.30& 78.77\\

& CDPruner
& 75.61 & 62.28 & 75.13 & 48.49 & 47.50 & 74.30 & 59.87 & 50.47 & 61.71 & 83.36 \\

& PruneSID
& 57.29 & 70.08 & 72.20 & 44.19 & 53.00 & 72.60 & 62.61 & 52.27 & 60.53 & 81.78 \\

& iLLaVA
& 71.01 & 50.68 & 72.69 & 51.89 & 53.40 & 75.23 & 63.14 & 53.33 & 61.42 & 82.98 \\

& ERASE
& 75.32 & 75.52 & 88.45 & 64.46 & 53.10 & 75.62 & 64.58 & 52.80 & \textbf{68.73}& \textbf{92.85}\\

\bottomrule
\end{tabular}
}
\end{table*}

\begin{table*}[t]
\centering
\caption{Results on Qwen3-VL-4B.}
\label{tab:qwen3_vl_4b}
\resizebox{\textwidth}{!}{
\begin{tabular}{l l|c c c c c c c c|c c}
\toprule
\textbf{Retain$\downarrow$}
& \textbf{Method}
& \textbf{TextVQA}
& \textbf{ChartQA}
& \textbf{DocVQA}
& \textbf{InfoVQA}
& \textbf{MathVista}
& \textbf{MMB\textsubscript{v1.1}}
& \textbf{RWQA}
& \textbf{MMStar}
& \textbf{Avg.$\uparrow$}
& \textbf{Rel. (\%)$\uparrow$} \\
\midrule

100\% & Base
& 81.56 & 82.68 & 94.79 & 79.94 & 65.30 & 82.89 & 71.24 & 65.53 & 77.99 & 100\\
\midrule
47\% & PDrop
& 70.23 & 55.76 & 80.71 & 55.32 & 46.60 & 74.77 & 67.06 & 52.07 & 62.82 & 80.54 \\
\midrule
\multirow{6}{*}{45\%}
& FastV
& 78.96 & 61.64 & 85.57 & 62.87 & 46.30 & 76.32 & 66.93 & 55.20 & 66.72 & 85.55 \\
& DART
& 71.54 & 73.00 & 83.08 & 57.84 & 56.10 & 79.41& 68.89& 56.73& 68.32& 87.60\\
& CDPruner
& 76.45 & 68.00 & 78.88 & 59.73 & 46.90 & 78.02 & 66.53 & 56.40 & 66.36 & 85.09 \\
& PruneSID
& 74.51 & 76.80 & 92.44 & 70.92 & 53.80 & 77.48 & 70.59 & 57.67 & \underline{71.78}& \underline{92.03}\\
& iLLaVA
& 76.38 & 73.00 & 91.42 & 71.31 & 48.40 & 78.72 & 72.68 & 58.47 & 71.30 & 91.42 \\
& ERASE
& 79.74 & 77.92 & 94.09 & 75.78 & 54.50 & 80.26 & 70.07& 61.27 & \textbf{74.20}& \textbf{95.14}\\
\midrule
\multirow{6}{*}{35\%}
& FastV
& 75.24 & 50.64 & 76.13 & 53.80 & 42.50 & 71.36 & 64.31 & 47.60 & 60.20 & 77.18 \\
& DART
& 65.67 & 64.80 & 73.55 & 48.69 & 50.40 & 76.93 & 66.67 & 54.07 & 62.60 & 80.26 \\
& CDPruner
& 73.19 & 60.04 & 70.81 & 53.62 & 43.20 & 75.62 & 66.27 & 50.80 & 61.69 & 79.10 \\
& PruneSID
& 69.39 & 68.92 & 89.52 & 64.03 & 50.10 & 74.69 & 69.02 & 54.53 & 67.53 & 86.58 \\
& iLLaVA
& 73.79 & 60.16 & 89.72 & 66.63 & 45.70 & 78.17 & 70.33 & 57.67 & \underline{67.77}& \underline{86.90}\\
& ERASE
& 78.36 & 74.28 & 93.38 & 71.59 & 52.90 & 79.49 & 68.24 & 59.80 & \textbf{72.26}& \textbf{92.65}\\
\midrule
\multirow{6}{*}{25\%}
& FastV
& 67.60 & 34.60 & 61.34 & 41.05 & 37.70 & 62.69 & 58.82 & 43.60 & 50.93 & 65.30 \\
& DART
& 57.79 & 53.72 & 57.84 & 37.48 & 45.30 & 74.54& 64.18& 48.60& 54.93& 70.43\\
& CDPruner
& 68.01 & 50.12 & 59.97 & 47.12 & 41.00 & 73.53 & 62.88 & 48.00 & 56.33 & 72.22 \\
& PruneSID
& 64.77 & 61.20 & 84.31 & 55.46 & 45.90 & 71.52 & 66.41 & 51.13 & 62.59 & 80.25 \\
& iLLaVA
& 72.84 & 52.96 & 87.96 & 54.44 & 45.20 & 77.17 & 69.93 & 56.13 & \underline{64.58}& \underline{82.80}\\
& ERASE
& 76.48 & 67.76 & 90.66 & 63.59 & 47.50 & 76.47 & 67.71 & 55.07 & \textbf{68.16}& \textbf{87.39}\\
\bottomrule
\end{tabular}
}
\end{table*}
\subsubsection{Evaluation on different model sizes}

ERASE achieves the highest average accuracy at all three matched retention ratios on Qwen2.5-VL-3B and Qwen3-VL-4B, supporting robustness across model sizes (Tables~\ref{tab:qwen2_5_3b} and~\ref{tab:qwen3_vl_4b}). On Qwen2.5-VL-3B, its gains over the strongest baseline are 0.66, 2.39, and 6.31 points at 45\%, 35\%, and 25\% retention, respectively. At 25\% retention, ERASE preserves 92.85\% of the base-model accuracy, compared with 84.33\% for FastV. On Qwen3-VL-4B, the corresponding average-accuracy gains are 2.42, 4.49, and 3.58 points. 

FastV and PDrop primarily use instruction relevance, while PruneSID relies on visual information. CDPruner and DART combine both sources but select tokens at a single point. iLLaVA distributes compression across stages, yet uses fixed stage-wise retention settings. ERASE instead jointly determines early token selection and image-dependent retention from raw-image statistics, then adjusts subsequent pruning to the resulting retention. Shallow pruning reuses these statistics alongside instruction relevance to preserve visual context, while deep pruning focuses on instruction relevance alone. These results demonstrate ERASE's effectiveness across diverse VLM architectures and model sizes.

\begin{table*}[b]
\centering
\caption{Accuracy and peak GPU memory on Qwen2.5-VL-7B.}
\label{tab:qwen2_5_7b_rescale}
\resizebox{\textwidth}{!}{
\begin{tabular}{l l|c c c c c c c c|c c}
\toprule
\textbf{Retain$\downarrow$}
& \textbf{Method}
& \textbf{TextVQA}
& \textbf{ChartQA}
& \textbf{DocVQA}
& \textbf{InfoVQA}
& \textbf{MathVista}
& \textbf{MMB\textsubscript{v1.1}}
& \textbf{RWQA}
& \textbf{MMStar}
& \textbf{Avg.$\uparrow$}
& \textbf{Peak. (GB)$\downarrow$} \\
\midrule

100\% & Base
& 85.30 & 86.00 & 94.82 & 82.31 & 65.10 & 82.51 & 70.20 & 65.53 & 78.97 & 19.37\\
\midrule

\multirow{4}{*}{45\%}
& AgilePruner
& 83.60& 84.56& 94.15& 75.40& 63.80& 81.58& 67.32& 64.20& 76.83& 38.81\\
& VScan
& 84.84& 83.76& 93.66& 78.02& 56.20& 78.02& 65.23& 58.93& 74.83& 37.38\\
& Nüwa
& 83.75& 84.60& 94.36& 77.51& 65.60& 81.58& 69.54& 63.73& 77.58& 39.59\\
& ERASE
& 84.52 & 83.24 & 94.22 & 79.27 & 64.20 & 81.27 & 70.20 & 64.13 & \textbf{77.63} & \textbf{19.15}\\
\midrule
\multirow{4}{*}{35\%}
& AgilePruner
& 82.26& 80.68& 92.27& 70.19& 63.10& 80.65& 67.45& 62.87& 74.93& 38.81
\\
& VScan
& 84.05& 82.64& 92.83& 74.35& 56.00& 77.63& 66.27& 57.40& 73.90& 37.38
\\
& Nüwa
& 82.82& 83.60& 93.68& 75.02& 64.50& 81.04& 70.07& 62.73& 76.68& 39.59
\\
& ERASE
& 84.26 & 83.04 & 94.12 & 76.89 & 63.70 & 81.66 & 69.80 & 63.93 & \textbf{77.18} & \textbf{19.15}\\
\midrule
\multirow{4}{*}{25\%}
& AgilePruner
& 80.12& 78.20& 88.10& 64.42& 57.60& 80.80& 67.19& 60.93& 72.17& 38.81
\\
& VScan
& 83.23& 79.60& 89.64& 68.29& 53.70& 76.78& 64.18& 55.13& 71.32& 37.38
\\
& Nüwa
& 80.51& 81.60& 92.22& 70.04& 61.40& 81.27& 69.80& 61.93& 74.85& 39.59
\\
& ERASE
& 83.50 & 80.20 & 93.35 & 71.83 & 62.20 & 81.50 & 69.41 & 62.60 & \textbf{75.57} & \textbf{19.15}\\
\bottomrule
\end{tabular}
}
\end{table*}

\subsubsection{Evaluation with downscaled inputs}\label{sec:appx-rescale}
AgilePruner, VScan, and Nüwa encounter OOM when processing inputs up to Qwen2.5-VL's default maximum of 16K visual tokens per image. Without an encoder \texttt{[CLS]} token for efficient attention-based scoring, their evaluated implementations require full attention maps, incurring substantial memory overhead. We therefore downscale their inputs to fit within the same GPU memory capacity.

Table~\ref{tab:qwen2_5_7b_rescale} presents these results. AgilePruner and Nüwa accommodate approximately 4.7K visual tokens per image, whereas VScan, which requires two attention maps, accommodates approximately 2.1K. Despite downscaling, their peak memory reaches 37.38--39.59~GB, roughly twice that of the base model. In contrast, ERASE achieves the highest average accuracy at every evaluated retention ratio using lightweight image statistics for early redundancy removal. It handles up to 16K visual tokens per image while using only 19.15~GB of peak memory, approximately half that of these baselines.

\subsubsection{Impact of $\tau$ for edge detection}
\begin{wrapfigure}{r}{0.3\textwidth} 
  \centering
  \vspace{-10pt}
  \includegraphics[width=\linewidth]{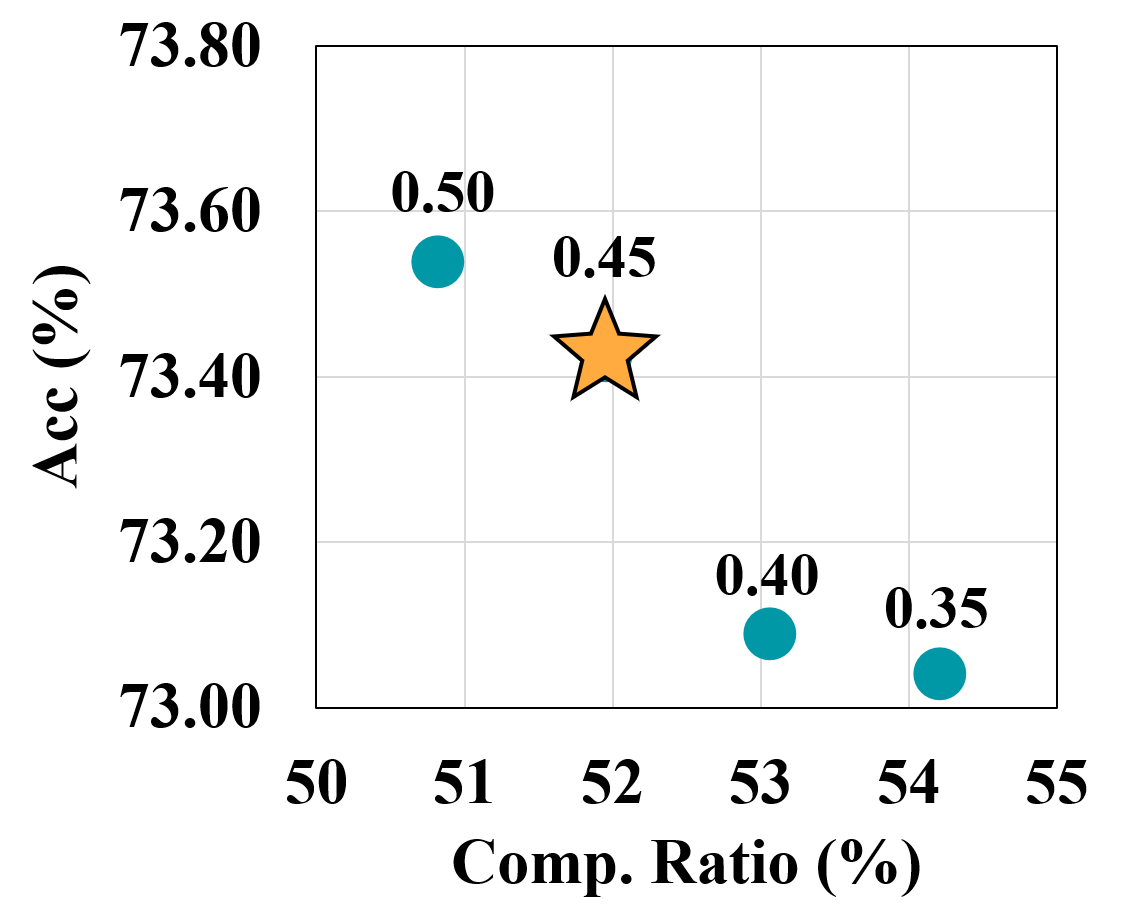} 
  \caption{
  Accuracy and compression trade-off for the edge-threshold parameter $\tau$.
  }
  \vspace{-10pt}
  \label{fig:appx-tau}
\end{wrapfigure}

As defined in Eq.~\ref{eq:edge_rule}, $\tau$ controls the threshold for edge detection. Increasing $\tau$ lowers the threshold and consequently retains more edge-containing patches, which generally improves accuracy at the cost of a lower compression ratio. To test the effect of this parameter, we evaluate $\tau\in\{0.35,0.40,0.45,0.50\}$ using only the edge-based criterion in Stage 1. We report the average accuracy and compression ratio over the four benchmarks used in our main ablation study: ChartQA, RealWorldQA, InfoVQA, and MMStar.

As shown in Fig.~\ref{fig:appx-tau}, $\tau=0.45$ provides a favorable operating point in the accuracy--compression trade-off. Increasing $\tau$ from 0.40 to 0.45 improves average accuracy by 0.34 points while reducing the compression ratio by 1.11 points. In contrast, a further increase from 0.45 to 0.50 reduces the compression ratio by a comparable 1.13 points but provides only a 0.11 point accuracy improvement. This diminishing return beyond $\tau=0.45$ motivates our choice of $\tau=0.45$ for all experiments.

\begin{table}[!b]
    \centering
    \caption{
        End-to-end component ablation on Qwen2.5-VL-7B.
    }
    \label{tab:component-ablation}
    \resizebox{0.85\textwidth}{!}{ 
    \begin{tabular}{cccc|cccc|c}
        \toprule
        Stage 1 & Shallow & VG & Deep
        & ChartQA & InfoVQA & RWQA & MMStar & Avg. \\
        \midrule
        $\times$ & $\checkmark$ & $\times$ & $\times$
        & 73.84 & 58.53 & 65.23 & 56.60 & 63.55 \\
        $\times$ & $\checkmark$ & $\checkmark$ & $\times$
        & 75.80 & 61.74 & 66.80 & 58.33 & 65.67 \\
        $\times$ & $\checkmark$ & $\times$ & $\checkmark$
        & 73.64 & 65.23 & 67.71 & 58.00 & 66.15 \\
        $\times$ & $\checkmark$ & $\checkmark$ & $\checkmark$
        & 76.64 & 67.23 & 67.71 & 61.87 & 68.36 \\
        \midrule
        $\checkmark$ & $\checkmark$ & $\times$ & $\times$
        & 79.68 & 67.22 & 67.97 & 60.00 & 68.72 \\
        $\checkmark$ & $\checkmark$ & $\checkmark$ & $\times$
        & 79.68 & 68.08 & 67.45 & 61.00 & 69.05 \\
        $\checkmark$ & $\checkmark$ & $\times$ & $\checkmark$
        & 79.48 & 71.48 & 68.24 & 61.20 & 70.10 \\
        $\checkmark$ & $\checkmark$ & $\checkmark$ & $\checkmark$
        & 80.20 & 71.83 & 69.41 & 62.60 & \textbf{71.01} \\
        \bottomrule
    \end{tabular}
    }
\end{table}
\subsubsection{End-to-end component ablation}
\label{sec:appx-component-ablation}

Complementing the analysis in Fig.~\ref{fig:stage2-c},
which measures retained accuracy on samples correctly answered
by the base model, Table~\ref{tab:component-ablation} reports
task accuracy on the full evaluation sets of the same four benchmarks.
All configurations use the same target layer-averaged visual
token retention of 25\%. VG denotes shallow-layer visual guidance, and $\checkmark$ and $\times$ indicate enabled and disabled components, respectively.
Shallow instruction-aware pruning is enabled throughout.
Disabling Stage 1 removes only its token pruning step, while the image statistics are still computed when visual guidance is enabled.

Stage 1 improves average accuracy across all four Stage 2
configurations. This improvement might be attributed to reduced attention
dilution among similar visual tokens, as Stage 1 removes
visually redundant tokens before instruction-aware selection.
Adding visual guidance also improves average accuracy both
with and without Stage 1, supporting the use of visual cues
alongside instruction relevance at the shallow layer to preserve useful
visual context.
Finally, adding deep instruction-aware pruning improves
average accuracy across all configurations under the same
overall token budget. This supports hierarchical selection
in Stage 2, where shallow pruning preserves visual context
and subsequent deep pruning further narrows the retained
tokens to instruction-relevant visual content.
The complete configuration achieves the highest average
accuracy of 71.01\%.

\begin{table}[t]
  \caption{
  Average efficiency across inputs with varying image resolutions.
  }
  \label{tab:appx-efficiency_metrics}
  \centering
    \resizebox{0.85\textwidth}{!}{ 
  \begin{tabular}{l| c c c c}
    \toprule
    \multirow{2}{*}{\textbf{Method}} & \textbf{KV Cache} & \textbf{Prefill Latency} & \textbf{Decode Latency}& \textbf{Task Latency} 
\\
    & (MB) $\downarrow$ & (ms) $\downarrow$ & (ms/token) $\downarrow$& (ms/task) $\downarrow$
\\
    \midrule
    Base & 409.43& 1990& 37.82& 2105\\
    \midrule
    FastV & 104.32 ($\times$3.92)& 1099 ($\times$1.81)& 34.17 ($\times$1.11)& 1206 ($\times$1.75)\\
    DART & 104.48 ($\times$3.92)& 1113 ($\times$1.79)& 35.37 ($\times$1.07)& 1219 ($\times$1.73)\\
    CDPruner & 104.33 ($\times$3.92)& 1560 ($\times$1.28)& 34.17 ($\times$1.11)& 1665 ($\times$1.26)\\
    PruneSID & 104.34 ($\times$3.92)& 1568 ($\times$1.27)& 33.15 ($\times$1.14)& 1671 ($\times$1.26)\\
 iLLaVA& 104.34 ($\times$3.92)& 1237 ($\times$1.61)& 33.45 ($\times$1.13)& 1341 ($\times$1.57)\\
    \textbf{ERASE (Ours)} & 104.32 ($\times$3.92)& 1099 ($\times$1.81)& 34.62 ($\times$1.09)& 1205 ($\times$1.75)\\
    \bottomrule
  \end{tabular}
    }
\end{table}
\subsubsection{Efficiency across image resolutions}\label{sec:appx-efficiency}
While Table~\ref{tab:efficiency_metrics} focuses on high-resolution inputs with 16K vision tokens, Table~\ref{tab:appx-efficiency_metrics} reports efficiency averaged over inputs spanning a broad range of image resolutions. At lower resolutions, the additional overhead of CDPruner and PruneSID becomes less dominant, resulting in higher average speedups than in the 16K-token setting. Nevertheless, both remain slower than other lightweight methods. CDPruner incurs repeated embedding-space subset selection, while PruneSID additionally materializes an $N\times N$ similarity matrix outside the accelerator. iLLaVA also introduces repeated token-merging operations across its compression stages. 
In contrast, FastV and DART achieve low latency through a single lightweight selection, limiting pruning overhead. Although ERASE applies multiple pruning stages, Stage 1 computes lightweight visual cues directly on a single-channel luminance image, and Stage 2 performs instruction-aware selection on this reduced token set. ERASE consequently achieves an average task latency of 1205~ms, comparable to FastV and DART. Together with the higher average accuracy reported in Table~\ref{tab:qwen2_5_7b}, these results show that ERASE's adaptive, hierarchical pruning preserves accuracy without sacrificing the latency advantage of lightweight selection.

\begin{figure}[!h]
  \centering
  \begin{subfigure}{0.44\textwidth}
    \centering
    \includegraphics[width=\textwidth]{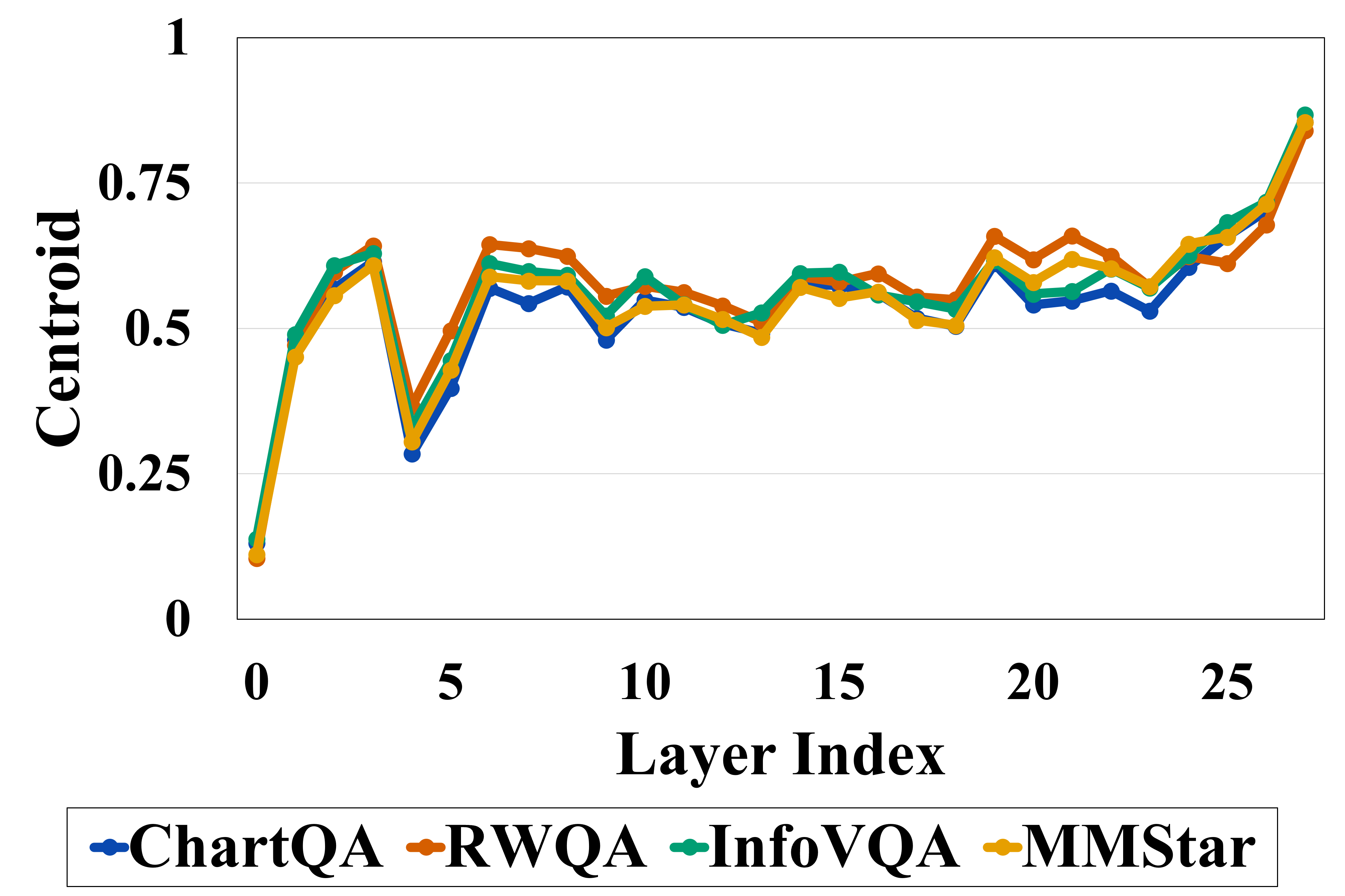}
    \caption{InternVL3-8B}
    \label{fig:app-stage2-a}
  \end{subfigure}
  \begin{subfigure}{0.44\textwidth}
    \centering
    \includegraphics[width=\textwidth]{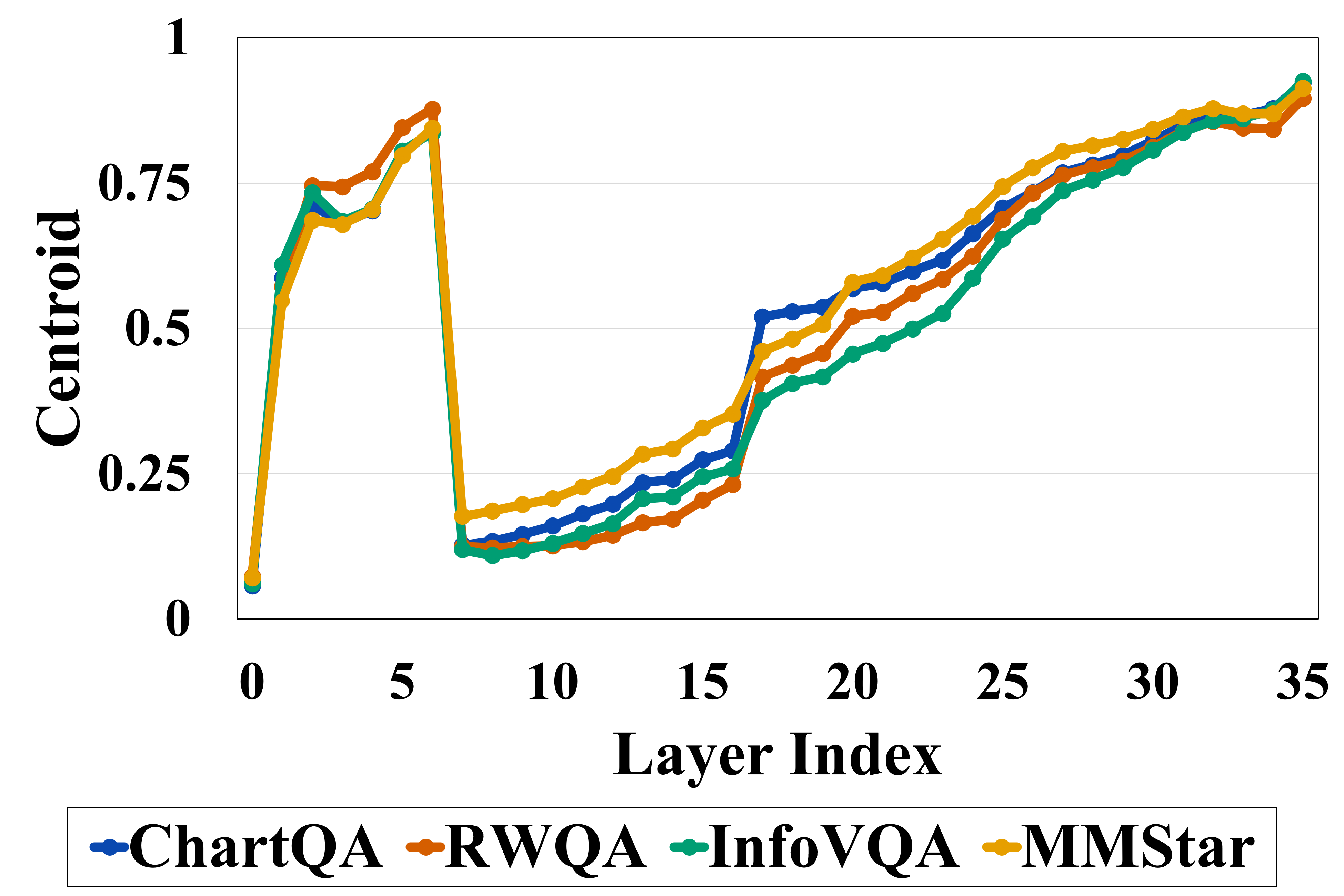}
    \caption{Qwen3-VL-8B}
    \label{fig:app-stage2-b}
  \end{subfigure}\hfill
  \begin{subfigure}{0.44\textwidth}
    \centering
    \includegraphics[width=\textwidth]{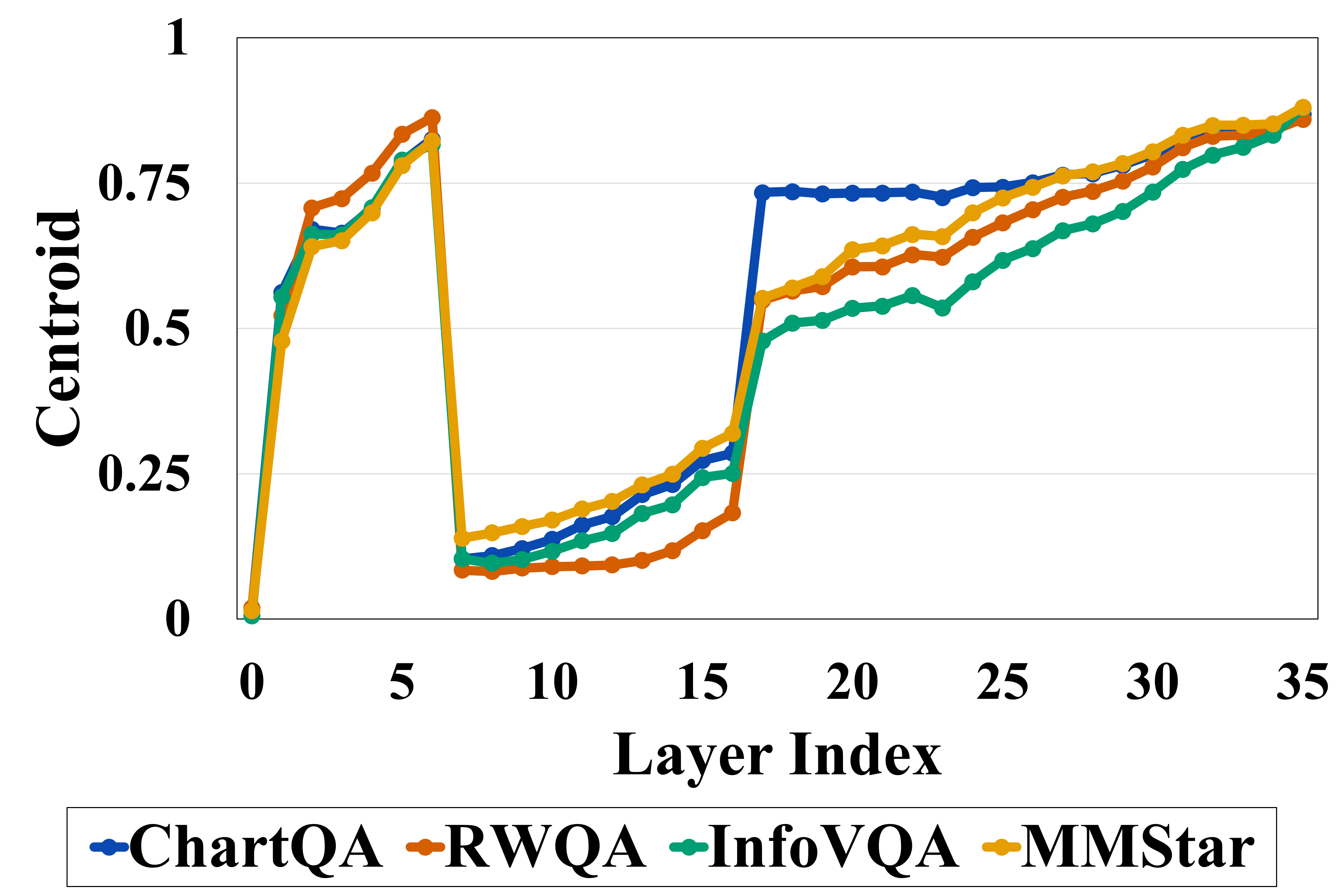}
    \caption{Qwen3-VL-4B}
    \label{fig:app-stage2-c}
  \end{subfigure}
  \begin{subfigure}{0.44\textwidth}
    \centering
    \includegraphics[width=\textwidth]{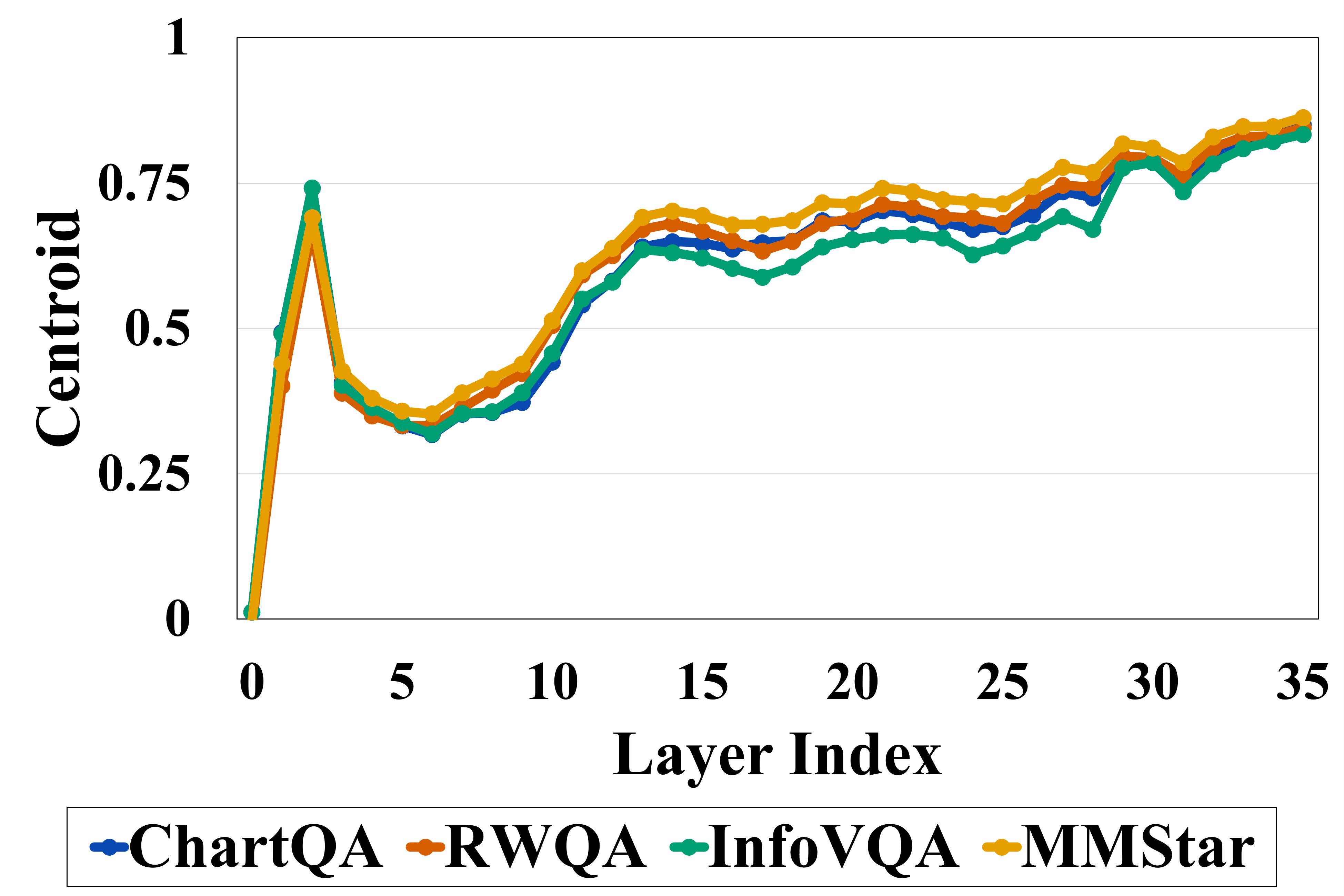} 
    \caption{Qwen2.5-VL-3B}
    \label{fig:app-stage2-d}
  \end{subfigure}
  \caption{Vision--text centroid similarity across decoder layers.}
  \label{fig:app-stage2}
\end{figure}

\subsection{Vision--text centroid similarity}\label{sec:appx-modality}
To examine whether the layer-wise pattern observed on Qwen2.5-VL-7B extends to other models, we analyze the vision--text centroid similarity defined in Eq.~\ref{eq:align_score} on InternVL3-8B, Qwen3-VL-8B, Qwen3-VL-4B, and Qwen2.5-VL-3B across four benchmarks (Fig.~\ref{fig:app-stage2}). Across these models, this alignment proxy exhibits a common pattern: an early rise, a subsequent drop, and recovery toward deeper layers, where its values are generally higher and more stable. Within each model, the benchmark curves closely follow one another, suggesting a largely consistent layer-wise pattern across the evaluated datasets.

Despite this shared pattern, the recovery trajectory varies across models. Qwen2.5-VL-7B reaches a mid-layer plateau with mild fluctuations before increasing further in deeper layers (Fig.~\ref{fig:stage2-a}). InternVL3-8B shows a fluctuating recovery, both Qwen3-VL models exhibit a distinct upward transition after the drop, and Qwen2.5-VL-3B recovers more gradually with some mid-layer fluctuations. These observations extend the qualitative pattern motivating shallow and deep pruning candidates in Section~\ref{sec:stage2-dual}, while showing that the recovery trajectory is model-specific.



\subsection{Visualization of retained tokens at each stage}
Fig.~\ref{fig:examples} visualizes the image regions corresponding to retained visual tokens after each pruning step. Stage 1 removes visually redundant regions while preserving informative structures throughout the image. The shallow and deep pruning steps of Stage 2 progressively focus the retained tokens on instruction-relevant content. Across these examples, answer-relevant details within the red boxes remain visible as surrounding regions are increasingly pruned. These qualitative results illustrate how the two stages jointly reduce visual redundancy and preserve information needed to answer the instruction.

\begin{figure}[!b]
  \centering
  \includegraphics[width=0.67\linewidth]{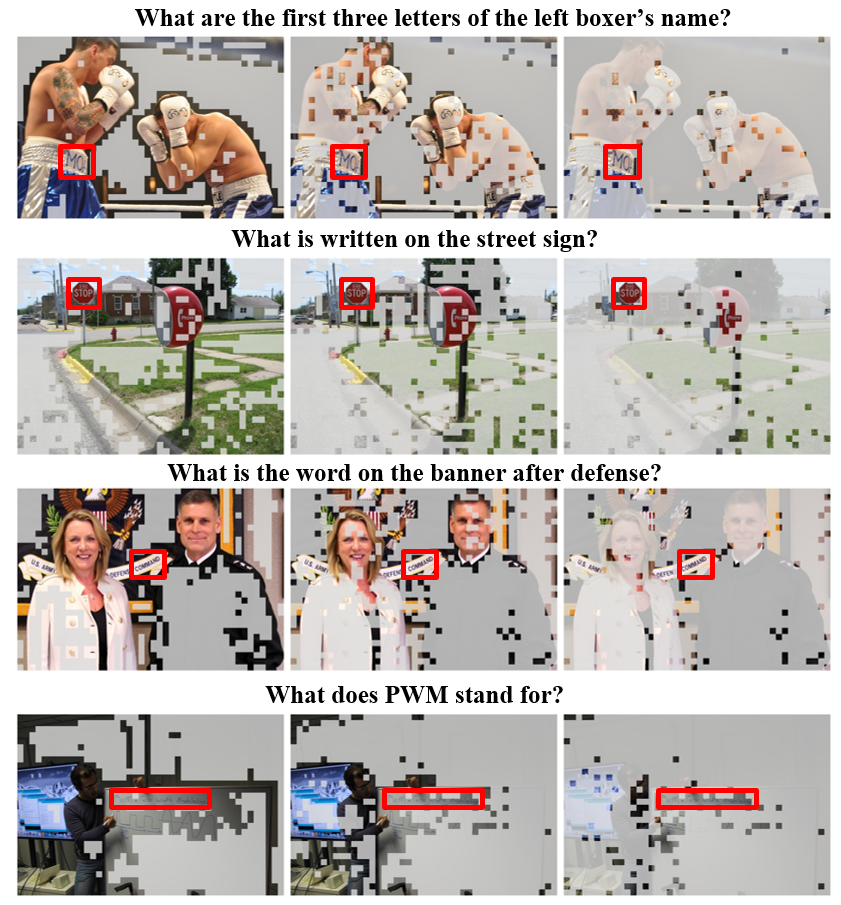}
  \caption{
    Retained visual tokens after Stage 1 (left), shallow Stage 2 pruning
    (middle), and deep Stage 2 pruning (right). Instructions appear above
    each row. Red boxes mark answer-relevant regions, and faded patches
    indicate pruned tokens.
  }
  \label{fig:examples}
\end{figure}

\clearpage

\begin{figure}[!t]
  \ContinuedFloat
  \centering
  \includegraphics[width=0.69\linewidth]{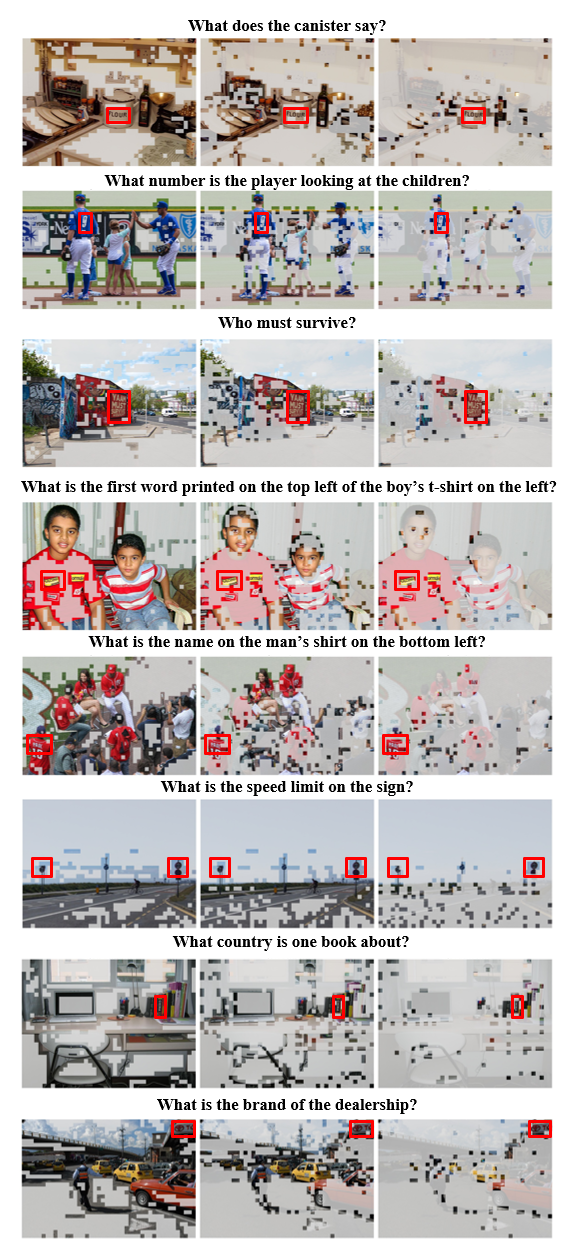}
  \caption{
    Retained visual token examples (continued).
  }
  \label{fig:examples-cont}
\end{figure}

\end{document}